\documentclass{clv2025}

\usepackage{lipsum}
\usepackage{amsmath,amsfonts,amsthm}
\usepackage{algorithmic}
\usepackage{algorithm}
\usepackage{array}
\usepackage[caption=false,font=normalsize,labelfont=sf,textfont=sf]{subfig}
\usepackage{textcomp}
\usepackage{stfloats}
\usepackage{url}
\usepackage{verbatim}
\usepackage{graphicx}
\usepackage{booktabs}
\usepackage{tikz}
\usepackage{xcolor}
\usepackage{float}   
\usepackage[dvipsnames]{xcolor}
\usepackage[most]{tcolorbox}
\usepackage{ragged2e}
\usepackage{xcolor}
\usepackage[utf8]{inputenc}
\usepackage{float}

\usepackage{mwe} 

\usepackage{amsmath}
\newcommand{\notimplies}{%
  \mathrel{{\ooalign{\hidewidth$\not\phantom{=}$\hidewidth\cr$\implies$}}}}

\title{Encoding and Propagation of Grammatical Number in Large Language Models}

\author{Jacob Solinsky$^{1}$ and Serguei Pakhomov$^{1}$}
\affilblock{\affil{$^{1}$University of Minnesota, USA\\
\textit{Corresponding author:} solin020@umn.edu}}

\usepackage{fancyhdr}

\begin{document}
\maketitle

\begin{abstract}
Sparse autoencoders (SAEs) have become an increasingly popular means of translating Large Language Model (LLM) hidden state vectors into a human-comprehensible list of features or symbols. As such, SAEs may provide insight into an LLM's knowledge representation and internal ``mental'' grammar - topics of pragmatic interest for mechanistic interpretation of AI behavior and of theoretical interest for the study of \textit{in silico} representations of human language. Language models with fewer than 100M parameters, trained on comparatively small corpora, nonetheless exhibit strong grammatical competence and discourse coherence, despite producing content of limited reliability. We use SAEs that we trained on one such model (Pythia-70M) to conduct an in-depth, comprehensive examination of how a single grammatical phenomenon, English number agreement, is represented and processed in an LLM.  We demonstrate the mechanism and the methodology by which SAEs can be used to find and propagate ``plural noun'', ``plural matrix subject'', and ``generate a plural verb'' features which cause the LLM to produce plural verb forms such as ``are'', ``do'', or ``have'' instead of ``is'', ``does'', and ``has'' in response to prompts such as ``the men in the park''. We show that intervening to remove these features from the hidden states of plural subject noun tokens causes the model to produce singular verbs instead, and vice versa. We also show that corresponding ``singular noun'' and ``generate a singular verb'' features are not as effective for this task. This asymmetry suggests that Pythia-70m makes use of a privative, rather than binary, representation of grammatical number in English. These findings replicated with larger LLMs Gemma-2-2b and Llama-3.1-8b yielded similar results. Our main contribution is a novel approach to the use of SAEs to understanding how LLMs encode human language that addresses a number of technical and methodological challenges.
\end{abstract}

\section{Introduction}
Large Language Models (LLMs) are a variety of Artificial Neural Network (ANN) which encode their "thoughts" in the language of high-dimensional vectors typically referred to in machine learning literature as the \emph{hidden states} that the network produces in response to a prompt. The field of mechanistic interpretability, also termed explainable Artificial Intelligence (xAI), aims to translate these indecipherable-by-humans hidden state-encoded thought processes into something a human researcher can understand.

The bulk of the continuing challenges facing LLM deployment,
be it the generation of hallucinations, or offensive and biased language,
or the production of insecure code, disgorging sensitive information
from a system prompt to users, or difficulty in staying on task,
have their roots in the fact that LLM hidden state is not transparent
to humans, unlike the debugging stack traces of conventional
symbolic computer programs written in a programming language. This lack of transparency prevents LLM developers and users from effectively debugging the LLMs. Instead of conventional debugging consisting of identifying and eliminating the root cause of an error directly and deterministically, debugging LLMs consists of indirect and probabilistic approaches such as additional rounds of Reinforcement Learning with Human Feedback (RLHF)  and/or prompt engineering. These strategies tend to address and fix specific instances of the LLM challenges rather than the underlying causes of these challenges, much like a medication such as acetaminophen may treat symptoms of influenza rather than act on the influenza virus itself. 

Despite these unresolved and ongoing challenges that resulted in limiting the use of LLMs in critical and sensitive areas including clinical and legal decision making, and scientific research and education, the one area in which LLMs unreservedly excel is language. Even smaller LLMs trained on limited amounts of text in a given language appear to be able to "learn" a complete syntactic theory of the language, as they rarely, if ever, generate ungrammatical sentences in that language even if the generated sentences may convey nonsensical or false information. For linguists, particularly those who have been working for decades in the generative syntax tradition, LLMs provide an unprecedented opportunity to test syntactic theories. More importantly, however, LLMs can also provide a means of discovering the system and the structure that underlies human language directly as opposed to making inferences based on behavioral observations. No linguist has yet created a complete theory of English grammar, or that of any other natural language, that could be used by a symbolic automaton to generate close to the full diversity of English sentences and larger narrative structures that are found throughout all of literature and the internet.

Although it is almost certain that LLMs process language in a completely different way than the human brain, given their dramatically different architecture and disembodiment, extracting a theory of natural language syntax from them that at least \emph{works} to generate a high-quality imitation of human language is certain to reinvigorate the study of theoretical syntax.\cite{piantadosi2023modern}. Given that LLMs are not transparent to humans, to fully realize this potential for linguistic discovery, there is a need to develop validated and reliable methods for interrogating and manipulating LLMs in a systematic and reproducible fashion.

The present paper contributes a step towards the objective of developing methods for extracting information pertaining to the structure that LLMs learn during pre-training. We demonstrate the utility of these methods by presenting an in-depth analysis of how an open weights transformer-based LLM (Pythia-70B) encodes the knowledge of subject-verb agreement in English, and confirm the generalizability of these findings on two larger LLMs (Gemma-2-2b and Llama-3.1-8b).   

\section{Background}

Early attempts to interpret LLMs struggled with what has been termed the \emph{polysemanticity problem} \cite{cunningham2023sparseautoencodershighlyinterpretable}. The analogies initially used to understand hidden states such as "vectors" drawn from physics, and "neurons" drawn from biology, proved a hindrance to understanding their true nature \cite{dey2025neuronbasedexplanationsneuralnetworks}. In physics, vectors capture concrete concepts of direction or position with the positions of values in these vectors having human-interpretable meanings, such as the classic cartesian 3-element $x,y,z$ vectors, often with the first element meaning 'distance along the horizontal', the second, `vertical', and the third, `depth', axes. This unwarranted assumption that ``positions in the hidden state vectors mean something'' was imported into early mechanistic LLM interpretability research. The biomimetic metaphors used in computational neuroscience reinforced this interpretability assumption by introducing the Artificial Neural Network (ANN) consisting of a collection of interconnected artificial neurons as a model for natural neural networks \cite{rumelhart_pdp_1986,churchland_computational_2016}. Each row of weights inside the feed-forward ANN matrices were thought to correspond to a set of "artificial dendrites" magnifying or diminishing the effect of a certain input on the neuron, the value resulting from the product of this row with the input vector being an ``activation potential'' signal being sent down that row's ``artificial axon''. Since biological neurons are real objects that can be directly measured, studied and manipulated, by analogy, this property of objecthood was erroneously bestowed on the concept of ``artificial neuron''. It has since been demonstrated that this analogy between ANNs and biological neural networks is not valid. The action potential traveling down a biological axon when the neuron "fires" cannot be modeled as a simple scalar. Instead, the action potential is more like a complex message encoded in a variety of schemes depending on the neuron type \cite{spruston_pyramidal_2008, mellem_action_2008}. Neuronal learning, even more poorly understood, is also a far more varied process than its artificial backpropagation counterpart used in ANNs \cite{wright_distinct_2025}. As such, it can take 1,000 artificial neurons to simulate a single biological one \cite{beniaguev_single_2021} which likely accounts for our persistent inability to simulate the 302-neuron \emph{C. elegans} nervous system \cite{evans_worm_2025} despite the development of computational hardware which can handle current frontier LLMs, rumored employ trillions of parameters \cite{palyekar_bytedance_2026}.

\subsection{Polysemanticity}

Despite the apparent lack of biological validity, this "positions mean something" assumption has been widely utilized in investigations of ANN behavior. The findings of such neuron interpretation attempts have generally been confusing. The results of \cite{dreyer_pure_2024}, provide a typical example. In their studies of the neurons of the penultimate layer  of the image classifier model ResNet50 (whose architecture is quite similar to that of an LLM), they found neurons such as this: Neuron \#1929 was highly activated on images of lipstick and wild boars. Neuron \#1506, on the other hand, was firing when presented images of clock faces, tall ground-walking birds, and combination locks. \#696 responded to hedge-mazes and toy-sized terrier dogs.  
This observation reflects the phenomenon termed \emph{polysemanticity}, a single neuron responding (or meaning) multiple things, which is a consequence of a basic aspect of linear algebra - the triviality of the change-of-basis transform. ANNs that consist of a series of matrices with each matrix representing a linear transformation with an \emph{orthonormal basis} can actually encode meaning in specific positions within the space spanned by their basis; however, these matrices can be easily converted into a system lacking this property, so long as there are not nonlinearities, such as activation functions, present in the system. For example, suppose we have an ANN system operating in a 3-dimensional vector space that encodes in its hidden state vectors the concept "noun" as $\langle 1,0,0 \rangle$, "singular" as $\langle 0,1,0 \rangle$, and "animal" as $\langle 0,0,1 \rangle$. This system can be transformed into one which encodes these same three concepts as $\langle 0.5,-3,2 \rangle$, $\langle 1.5,7,4 \rangle$, and $\langle 6,-2.5,1 \rangle$ without changing any of its externally visible behavior \cite{elhage2022toymodelssuperposition}. In other words, the destination point does not change with the change of basis - but now there are two ways to reach that destination using different systems of coordinates (i.e., \emph{polysemanticity}). The incorporation of nonlinear activation functions into an ANN, like the GeLU used in an LLMs MultiLayer-Perceptron layer, do somewhat encourage ANNs to  use an orthonormal basis, thus somewhat concentrating meaning into individual ``artifical neurons'', but this caveat still does not eliminate all polysemantic neurons \cite{elhage2022toymodelssuperposition}.

\subsection{Sparse Auto-Encoders}

Sparse autoencoders (SAE) are designed to help untangle polysemanticity. Their inventors held fast to the assumption that  ANNs do encode meaning in their hidden states using a human-interpretable system of basis vectors. The only assumption they relaxed was that this basis is orthonormal. The next step in deciphering this system is to devise a means of uncovering the basis. A variety of approaches exist to do this, as it is not well defined mathematically whether the basis the SAE has learned really captures the ``semantic atoms'' of the internal language the LLM uses. Whether the SAE has accomplished this well is unfortunately a matter of subjective human judgment.

SAEs transform an input hidden-state vector of dimension $\mathit{D}$ into a \emph{feature vector} in a much larger space, whose elements are restricted to being >=0. The SAEs discussed in this paper use a feature space of size $\mathit{D} * 32$. Because the feature space is so much larger than the original hidden state space, a naively designed non-sparse autoencoder would learn an uninteresting, mathematically degenerate encoding, by devoting the first $\mathit{N}$ positions in feature space to each dimension of the orthonormal basis of the hidden space, and then zeroing out everything else. The property of "sparseness", i.e., the feature vectors being restricted to having a small number (typically a few dozen) nonzero elements, is added to the autoencoder to prevent this degenerate failure mode.

Unlike hidden state vectors, SAE feature vectors drawn from LLM hidden states do generally have that desirable "each position in the vector means something" property, termed \emph{monosemanticity}. For example, a team at Anthropic trained a 34-million-dimensional SAE on the hidden state vectors of Claude 3 Sonnet LLM to discover that nonzero values at position \#31164353 were often found in passages discussing the Golden Gate bridge or that position \#9493533 was associated with passages pertaining to brain sciences, among many other such associations \cite{templeton_scaling_2026}. Following this work by Anthropic, we propose that an SAE can be viewed as translating an ANN's hidden state vectors into short lists of symbols. While such a symbolic representation is much more comfortable and familiar to scientists than working with LLM parameters, the SAE symbols still present a challenge because they belong to an unknown and ``alien" language. A typical SAE generates hundreds of thousands to millions of features/symbols in the course of its training on an LLM. Authoring a database, or dictionary, of what each symbol means is a task more monumental than that faced by the 19th-century Assyriologists who first deciphered the meanings of the hundreds of cuneiform symbols of Akkadian and Sumerian \cite{hasselbachandee_decipherment_2020}.

\subsubsection{Framework for examining "meaning" of SAE features}

In the above account of Anthropic's work, we must be careful to be rigorous about what is meant by an SAE feature ``meaning'' something, taking heed of the fact that LLMs do not have senses or a notion of time and, although they are made of language, they do not use language as we do. Hence we chose a behaviorist framework to explore ``meaning" captured by SAE features. In this framework, the cause or stimulus of all LLM behavior is the prompt propagated through the LLM weights, and the effect or observable behavior consists of the tokens produced in response to the prompt. The hidden state vectors, and the SAE features which comprise them, are but intermediaries in these causal chains, and derive their meaning from the causal chains in which they participate. We can then speak, however loosely, of ``meaning'' in a more human sense by how the patterns of these relationships between prompt, features, and generated output mean something when read by an English speaking person, even though it is perfectly possible to create an LLM which maps from gibberish prompt to gibberish output according to logic inscrutable to an outside observer.

We use Pearl's causality framework \cite{pearl_causality_2000} to organize our methods for quantifying the ``meanings'' of SAE features. We map the prompt given to the LLM as forming the exogenous variables $U$. As prompts are generally written in English or other natural language, we can use the denotative semantics of natural languages to anchor the variables in $U$ to more conventional semantic frameworks. $V$, the endogenous variables, consist of both the SAE features embedded in the model's hidden state vectors and the tokens it ultimately generates. We subdivide $V$ into $F$, the SAE features appearing on the model's hidden states, and the output token probability distributions which are actually used for token generation, for which we designed a function $B$ that approximates the notion of ``the likelihood that the model will produce a singular or plural verb''. $E$ constitutes the causal edges between the variables in $U$ and $V$, forming what are here termed ``circuits'' between prompt tokens in $U$, the features of $F$ and changes in the output token generation probability distributions quantified by $B$. Our framework prioritizes \textit{model behavior}. As will be discussed later, we found that Pythia-70m \textit{incidentally marks} more features on plural nouns than actually \textit{have an effect} on verb conjugation when they are altered.

\subsection{Attention Mechanism}

The introduction of the attention mechanism to the Transformer ANN architecture in language modeling in 2017 resulted in a qualitative leap forward towards what we now know as LLMs that have become synonymous with Artificial Intelligence. The details of the self-attention mechanism are numerous and available in the original paper that introduced it \cite{vaswani2023attentionneed}. Briefly, self-attention consists of a mechanism that allows the ANN to "notice" and "remember" which tokens in the context of the target token that the ANN is trying to predict at any given point in time are associated with the correct choice. This mechanism is composed of three matrices $W_K$ (Key), $W_Q$ (Query) and $W_V$ (Value). Each token in the input (represented as its embedding vector) is multiplied with the $W_K$, $W_Q$, and $W_V$ matrices to create a set of ${\color{OliveGreen}q}$, ${\color{blue}k}$, and ${\color{red}v}$ vectors. The ${\color{OliveGreen}q}$ and ${\color{blue}k}$ vectors are multiplied to create a table of attention scores, which helps the model figure out which tokens have a lot to do with each other and which  have little relationship. For example, in the sentence ``We were sitting on the river bank'', the ``river'' and ``bank'' tokens would have a high attention score between them, helping the model use the context of ``river'' to disambiguate the geographical and financial meanings of ``bank''. The attention scores are multiplied with the value vectors to give the attention mechanism's output. In the preceding example, the token ``bank'' will have an output containing a high proportion of the value vector of ``river'' because of the high attention score between them, while having its output not much influenced by the token ``were'' to which less attention was paid. An LLM based on the Transformer architecture with self-attention has multiple self-attention mechanisms stacked into  ``layers'', each one able to independently learn different attention patterns appropriate to the stage of language processing it is responsible for depending on how deeply it is positioned within the model.

\subsection{SAE Inference and Training Algorithms}

A sparse autoencoder of expansion factor $f$ maps from LLM hidden states having $d$ \textit{hidden dimensions} to a feature space having $f * d$ dimensions and back again. It consists of an encoder matrix $W_e \in \mathbb{R}^{d \times f \cdot d}$ with associated encoder bias $b_e \in \mathbb{R}^d$, and a symmetrical decoder weight and bias, $W_d \in \mathbb{R}^{f \cdot d \times d}$ and $b_d \in \mathbb{R}^d$. These two sets of parameters are occasionally tied, such that $b_d = b_e$ and $W_d = W_e^\top$; however, these symmetrical sets of parameters are more often allowed to be learned independently. Two common methods are used to enforce sparseness in SAEs, the one we used, the TopK function which zeroes out all but the $k$ highest elements of the feature vector it outputs, a method preferred by OpenAI's researchers \cite{gao_scaling_2024} and the $L_1$ penalty SAE, which adds a taxicab metric loss to the loss calculations to encourage the SAE to zero out the majority of the elements in its feature vectors during training, a tactic preferred by Anthropic's transformer-circuits.pub team.

The feature vector $z$ is generated from input $h \in \mathbb{R}^n$ with the equation $z = W_e h + b_e$

If a TopK SAE is being trained, sparsity will be enforced with the TopK
function, which zeroes out all but the k-highest elements of $z$: $z = \mathrm{TopK}(W_e h + b_e, k)$

And the reconstructed vector $\hat{h}$ is calculated in this way: $ \hat{h} = W_d z - b_d$

During training, loss is calculated as the mean squared error between $h$ and $\hat{h}$:

\[ \mathcal{L} = \|h-\hat{h}\|_2^2\]

Because the TopK function prevents backpropagation from reaching any of 
the low-strength features that were excluded by it, during training,
auxiliary loss $\mathcal{L}_{aux}$ is periodically added to the regular loss by calculating an auxiliary reconstruction vector $\hat{h}_{aux}$ using exclusively ``dead'' features which have not been selected by TopK for a while, an operation
denoted with the function $\mathrm{DeadMask}$

\begin{gather*}
z_{aux} = \mathrm{DeadMask}(W_e h + b_e) \\
\hat{h}_{aux} = W_d z_{aux} - b_d \\
 \mathcal{L}_{aux} = \|h-\hat{h}_{aux}\|_2^2
\end{gather*}

We will subsequently refer to an SAE `` decomposing'' an input hidden state vector $h$ into either ``features'' (the term used in SAE literature) or ``symbols'' (emphasizing the notion of these vector decompositions being denotative symbols in the semantic sense). We then say that ``symbol $S$ is inside hidden state vector $h$''. Because all hidden state vectors belong to a specific token in the LLM's prompt, we can then make metonymic statements such as ``The LLM places symbol $S$ (which perhaps denotes the grammatical singular) on the token ` deer' given the prompt `one deer' as a result of (some calculation)''. This abbreviated language will assist in not overburdening the reader with contextually irrelevant details such as ``Our SAE for MLP sublayer 3 found feature $S$ at a strength of 1.34, which we considered to be significant, within hidden state vector ${\color{brown}m_2^3}$ inside the prompt `One deer', of which the token ` deer' is the second token.''

Tables \ref{tab:dimensions}, \ref{tab:states}, and \ref{tab:dimensions} in the appendix contain a comprehensive overview of all mathematical symbols we use in this paper.

\subsubsection{Pythonic SAE Feature notation}
We interchangeably use two forms of notation for SAE features.
In Pythonic notation, the value of the 430th feature of the attention subcomponent of the 5th layer of Pythia-70m as it appears on the 3rd token (``on'') in the prompt $s$ ``The man on the bench'' is 4:attn:429@2 in $s$. This is equivalent to, in our mathematical notation: $\text{model}(s).{\color{teal}O_3^5} \cdot {\color{teal}\text{SAE}{(O^5)_{430}}}$, the dot product of the appropriate $\color{teal}O$ vector with the 430th decoder feature vector in the matching SAE. Python is a 0-indexed language, unlike standard mathematical notation, which is why we found this notation easier to work with in our code. Our notation for model subcompoments outputs in python was
\begin{itemize}
\item: Embedding matrix's $f$th feature =we:$f$-1 = $\text{SAE}{(W_E)_{f}}$
\item: Unembedding matrix's $f$th feature = wu:$f$-1 =$\text{SAE}{(W_U)_{f}}$
\item: $\ell$th Attention sublayer's $f$th feature =$\ell$-1:attn:$f$-1 = ${\color{teal}\text{SAE}{(O^\ell)_{f}}}$
\item: $\ell$th MLP sublayer's $f$th feature = $\ell$-1:mlp:$f$-1 =${\color{brown}\text{SAE}{(M^\ell)_{f}}}$
\item: $\ell$th value sublayer's $n$th head's $f$th feature = $\ell$-1:v$n$-1:$f$-1 = ${\color{red}\text{SAE}{(V^{(\ell,n)})_{f}}}$
\item: $\ell$th key sublayer's $n$th head's $f$th feature = $\ell$-1:k$n$-1:$f$-1 = 
${\color{blue}\text{SAE}{(K^{(\ell,n)})_{f}}}$
\item: $\ell$th query sublayer's $n$th head's $f$th feature = $\ell$-1:q$n$-1:$f$-1 =
${\color{OliveGreen}\text{SAE}{(Q^{(\ell,n)})_{f}}}$

\end{itemize}

\subsection{Pythia-70m language model architecture}

The Pythia family of models were created as an open-source resource for LLM researchers outside of proprietary corporate environments\cite{biderman2023pythiasuiteanalyzinglarge}. All Pythia LLMs use the same decoder-only Transformer architecture as other frontier commercial and non-commercial LLMs but differ in the number of parameters\footnote{with the caveat that this may not be true of the current state-of-the-art frontier commercial LLMs whose architecture is not publicly available.}. Certain aspects of its architecture were designed with aiding interpretability research in mind. In this section we provide a far more in-depth description of the operation of Pythia models than is typical in the literature. We believe this level of depth is necessary in this particular paper (even for readers familiar with Transformers and Pythia) to facilitate a precise understanding of the subsequent explanation of how we use SAEs to interpret Pythia's internal operations on hidden state vectors as symbolic calculations.

\begin{figure}[H]
  \centering
  \includegraphics[width=0.7\textwidth]{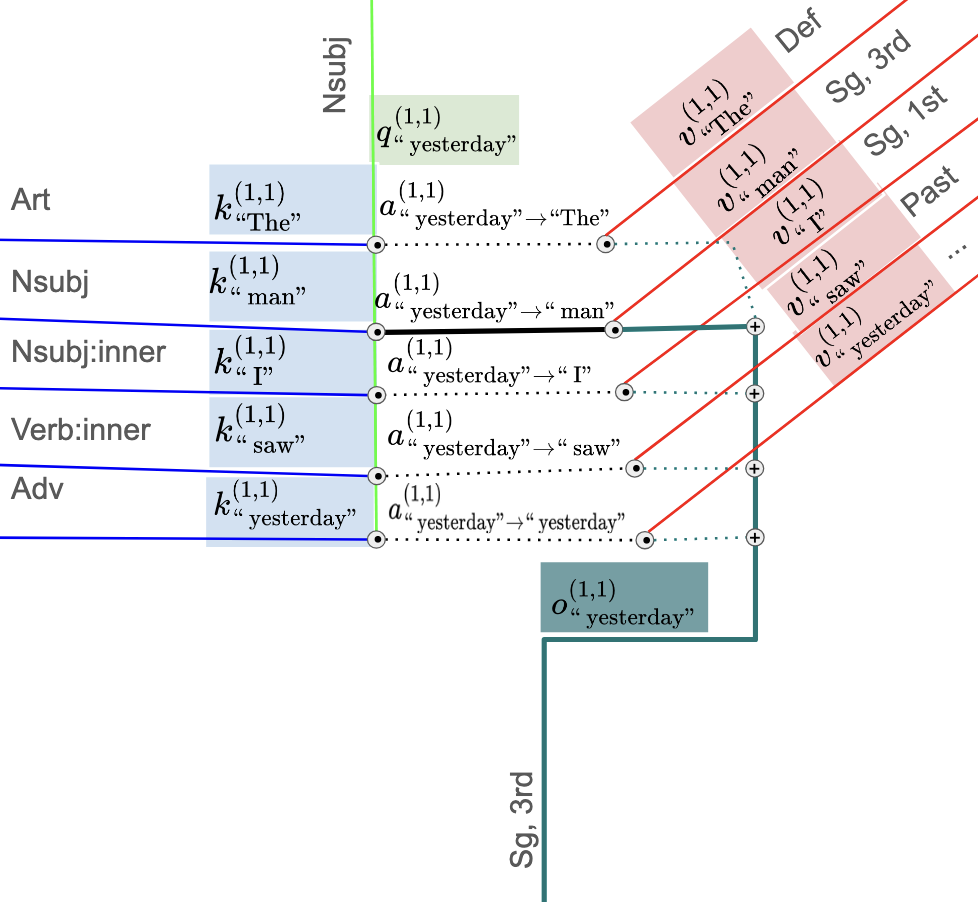}
  \caption{\textbf{Sg} is copied to the output vector}
  \label{fig:AttentionGrammar3}
\end{figure}

\begin{figure}[H]
  \centering
  \includegraphics[width=0.7\textwidth]{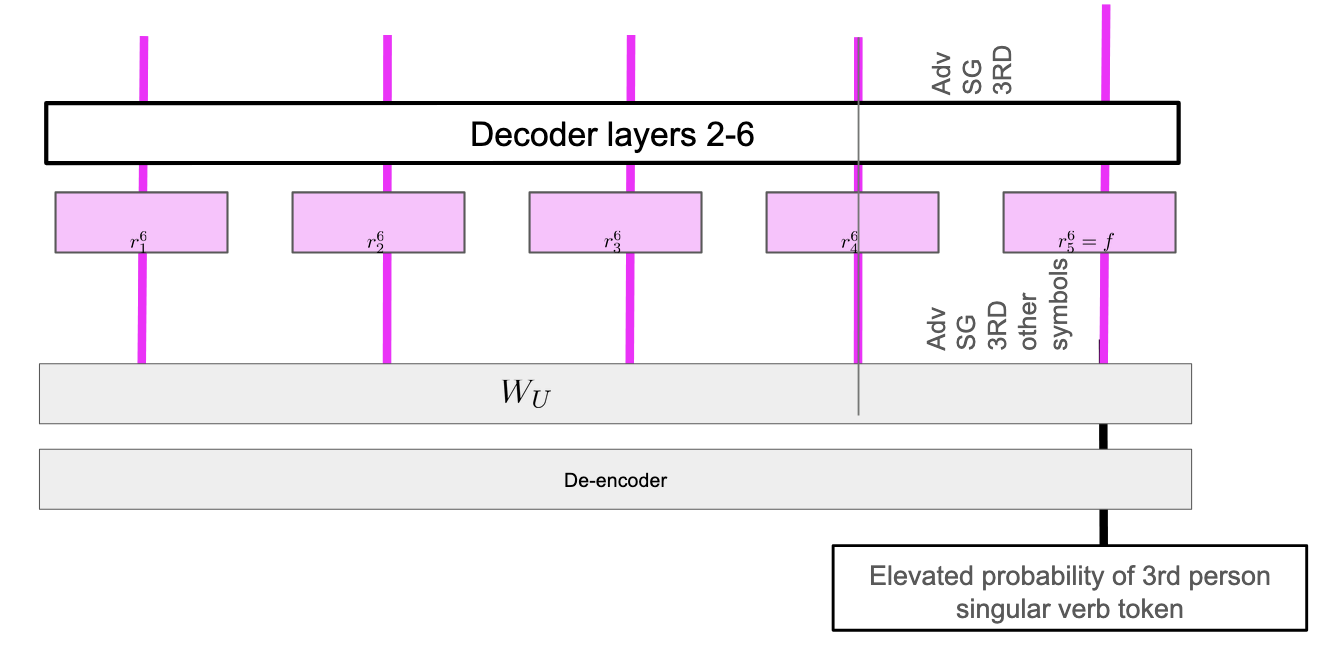}
  \caption{\textbf{Sg} symbol read by the $W_U$ matrix to output singular verbs}
  \label{fig:AttentionGrammar4}
\end{figure}

The decoder-stack model architecture used by Pythia is illustrated in detail in  circuit diagram-style figures we have included in the appendix (Figures~\ref{fig:LLMTopLevel} \ref{fig:LLMAttentionFocus}), which are numbered corresponding to the equations in the text  walkthrough. We include in the flow of the text figures that focus on the most important portions of them (Figures \ref{fig:Man3rdPersonSingula}, \ref{fig:AttentionGrammar2}, \ref{fig:AttentionGrammar3}, \ref{fig:AttentionGrammar4}). The diagrams use the following conventions. Brightly colored lines represent hidden state vectors. The notation label for each vector is placed in a box of a matching color that the vector's line passes through at some point along its trajectory. Symbols $\odot$ and $\oplus$ indicate the dot product and vector addition, respectively. The bullet symbol $\bullet$ denotes a vector ``branching'' when it is used in more than one place in the model, to help disambiguate from visual line-crossings which do not signify any mathematical operation. 

When we interpret a vector as carrying SAE symbols, we write said symbols parallel to the vector's colored line, as if they were messages being passed down a wire. When two vectors are added together, the child vector inherits all of the SAE symbols upstream of it. When the dot product is taken of two vectors, the output is proportional to how many symbols overlap between the two vectors and the strength at which those symbols are present. Transformations (for the most part matrix multiplications and bias additions) are indicated with large uncolored boxes. Normalization steps (both the RMS norm applied each time a vector is added to the residual stream and the softmax norm applied to attention scores to make them add up to one) are ignored, as neither has a strong effect on the symbolic content of the vectors passed through them.   The corresponding notation for the vector in all cases resembles the form  $\color{BrickRed}{h_j^{\text{coordinates}}}$ or $\textcolor{BrickRed}{h_\text{`` man''}^{\text{coordinates}}}$, where $\color{BrickRed}{j}$ is the jth token in the input to which the vector belongs (i.e., the word `` man'' in this example). Note that all vectors in decoder-stack LLMs belong to a specific token in the prompt, none float freely.  In certain cases $j$ is disambiguated to the token itself (e.g., \textcolor{BrickRed}{`` man''}).  The type of hidden state vector is denoted by a lowercase letter with a unique characteristic color to enable easier interpretation of the diagrams as follows:

\begin{itemize}
\item \textcolor{magenta}{magenta} for $\color{magenta}{r}$ residual stream vectors - foundational memory structures in Pythia that allow information to flow between different layers. They are continually written to via addition and are the only vectors which cross model component boundaries.
\item \textcolor{violet}{violet} for $\color{violet}{O}$ for attention sublayer output  vectors.
\item \textcolor{orange}{orange} for ${\color{orange}r_{j}^{{\ell-1}:attnnorm}}$ for attention sublayer normed input vectors

\item \textcolor{brown}{brown} for $\color{brown}{m}$ for MLP sublayer output vectors.
\item \textcolor{Peach}{peach} for ${\color{Peach}r_{j}^{{\ell-1}:mlpnorm}}$ for MLP sublayer normed input vectors
\item \textcolor{gray}{gray} for $\color{gray}{m^{\ell:i}}$ for MLP sublayer intermediate vectors.
\item \textcolor{OliveGreen}{green} for $\color{OliveGreen}{q}$ for query vectors
\item \textcolor{blue}{blue} for $\color{blue}{k}$ for key vectors.
\item \textcolor{red}{red} for $\color{red}{v}$ for value vectors.
\item \textcolor{teal}{teal} for $\color{teal}{O}$ for attention head outputs.
\end{itemize}
We have made our notation more thorough than was strictly necessary for describing our work here, as we anticipate that this notation will be useful both in our future works and possibly to other explainable AI (xAI) researchers.

\subsection{Residual streams}

In this section, we present a step by step walkthrough of Pythia's residual streams which carry information through the LLM layers.

\subsubsection{Tokenization}

Each Pythia model has a token vocabulary $V$. For Pythia-70m, the vocabulary has 50304 tokens. The embedding matrix, $W_E \in \mathbb{R}^{|V| \times d}$, translates tokens into embedding vectors of size $d$, called the ``hidden size''. These embedding vectors comprise the foundation of the model's hidden state vectors, which accumulate information as they are processed by the model's various encoder layers. A symmetrical unembedding matrix $W_U \in \mathbb{R}^{d \times |V|}$ exists at the end of the model to translate hidden state vectors back into tokens.

Presented with input, the model begins by dividing it into tokens, which will be whole words in the case of high-frequency items and word parts in other cases. Pythia, like most frontier LLMs today, makes no attempt to divide low-frequency or conjugated words into linguistically defined morphemes, meaning that subword elements like noun or verb suffixes are not processed as discrete entities by the tokenizer, a detail which will become very important for our purposes later. It also follows the tradition of most modern LLMs in incorporating leading spaces into its tokens. These tokens are then converted into token numbers, which are then looked up to find the corresponding embeddings vectors in $W_E$. For example, the token `The' is number 510 in Pythia's vocabulary, and the 510th vector of $W_E$ becomes ${\color{magenta}r_1^0}$. These ${\color{magenta}r}$ vectors each form the source of what are called  \emph{residual streams} (RS). Each subsequent attention sublayer and MLP sublayer create \emph{contribution} vectors: $\color{purple}O_j^\ell$ and $\color{purple}m_j^\ell$ respectively,  which are added to, or ``flow into'', the input residual stream after each layer, like tributaries. These ${\color{teal}O_j^\ell}$ and ${\color{brown}m_j^\ell}$ vectors, the ``tributaries'', are what the SAE interprets, not the ${\color{magenta}r_j^\ell}$ residual stream vectors, which contain a harder-to-interpret mixture of information summed together from every part of the model upstream of them. When we want to refer to these tributary vectors collectively, we term them $h_j^\ell$. A potentially confusing point is that the ``hidden states'' returned by many LLM APIs like that of Hugging Face's \texttt{transformers} are the ${\color{magenta}r_j^\ell}$ vectors, not the $h_j^\ell$ vectors. The Gemma Scope SAE project \cite{lieberum_gemma_2024} trained SAEs for both the ${\color{magenta}r_j^\ell}$ and $h_j^\ell$  vectors.

\begin{gather*}
\text{``The man I saw yesterday''} \\
(\text{`The'},\text{` man'},\text{` I'},\text{` saw'},\text{` yesterday')} \\
(510,   637,   309,  3047, 11066) \\
({\color{magenta}r_1^0,r_2^0,r_3^0,r_4^0,r_5^0}) \\
{\color{magenta}r_j^{\ell}} = {\color{pink}r_j^{\ell-1}} + {\color{brown}m_j^\ell} + {\color{teal}O_j^\ell}
\\
\end{gather*}

There is one residual stream per token in the input that can be thought of metaphorically, in this case, as representing 6 parallel river tributary systems. Information "flows" down the tributary vectors into these streams and is merged together by addition with little loss as long as the tributary vectors (${\color{teal}O_j^\ell}$ and ${\color{teal}m_j^\ell}$) are mostly orthogonal to each other (an unintuitively much easier property to achieve in the high-dimensional spaces used by LLMs than a visualization using 3D vectors would suggest). Picture each stream as "carrying" a set of symbols (an SAE decomposes its input vectors into multiple symbols, not just once), and that the number of symbols increases as more of these streams merge together.

We now go into the details of how the tributary vectors, ${\color{teal}O_j^\ell}$ and ${\color{teal}m_j^\ell}$, are calculated. These two vectors come from the self-attention and MultiLayer Perceptron sublayers, respectively, a pair of which comprise a \textit{decoder layer}. There are 6 decoder layers in Pythia-70m. The MLP and self-attention sublayers of a layer are calculated in parallel in Pythia-70m, which is untrue of other model architectures like Gemma, in which MLP comes after self attention.  We will go over the more complex and interesting of these sublayers, the self attention mechanism, first.

\subsubsection{Attention Sublayer}

The attention sublayers are the only point in an LLM where information can cross over from one residual stream to another. An attention sublayer consists of several attention heads, $n_{head}$, which we index with the variable $n$. Each of these heads consists of a triplet of weights matrices and bias vectors
\begin{gather*}
W_Q^{(\ell,n)}, W_K^{(\ell,n)}, W_V^{(\ell,n)} \in \mathbb{R}^{d \times d_{head}} \\
b_Q^{(\ell,n)}, b_K^{(\ell,n)}, b_V^{(\ell,n)} \in \mathbb{R}^{d_{head}} \\
d_{head} = d /n_{head}
\end{gather*}

Each input residual stream $r_j^{\ell-1}$ to head $n$ of layer $ell$,  is used to 
generate a triplet of $d_{head}$ dimensional vectors,  the query, key, and value.
\begin{equation}
{\color{orange}r_{j}^{{\ell-1}:attnnorm}} = \frac{{\color{magenta}r_j^{\ell-1}} - \mathbb{E}[{\color{magenta}r_j^{\ell-1}}]}{
\sqrt{\text{Var}[{\color{magenta}r_j^{\ell-1}}] + \epsilon}
}*\gamma^\ell_{KQV} + \beta^\ell_{KQV}
\end{equation}

\begin{equation}
{\color{OliveGreen}q_j^{(\ell, n)}} = \text{RoPE}(W_Q^{\ell, n}{\color{orange}r_{j}^{{\ell-1}:attnnorm}} + b_Q^{\ell, n})
\end{equation}

\begin{equation}
{\color{blue}k_j^{(\ell, n)}} = \text{RoPE}(W_K^{\ell, n}{\color{orange}r_{j}^{{\ell-1}:attnnorm}} + b_K^{\ell, n})
\end{equation}

\begin{equation}
{\color{red}v_j^{(\ell, n)}} = W_V^{\ell, n}{\color{orange}r_{j}^{{\ell-1}:attnnorm}} + b_V^{\ell, n}
\end{equation}

After the RoPE function ``rotates'' pairs of elements in the $q$ and $k$, each head then calculates individual raw attention scores $p_j^{\ell,n,i}$ (combining into a sequence $P_j^{(\ell, n)}$) for each residual stream by taking the dot product of the residual stream's query vector $\color{OliveGreen}q_j^{\ell, n}$ with each $\color{blue}k_i^{\ell, n}$ with $i$ ranging from $1$ to $j$. Each residual stream only receives attention scores for any tokens to its \emph{left}, including itself, any scores to the right are zeroed out. 

\begin{equation}
p_j^{\ell,n,i} = 
\begin{cases}
{\color{OliveGreen}q_j^{(\ell, n)}} {\color{blue}k_i^{(\ell, n)}}, & i \leq j \\
0, & i>j \\
\end{cases}
\end{equation}

The raw attention scores $P_j^{(\ell, n)}$ are then normalized into probabilities summing up to 1, $A_j^{(\ell, n)}$, using the softmax function
\[A_j^{(\ell, n)} = \text{Softmax}(P_j^{(\ell, n)})\]

The output $\color{teal}o_j^{\ell,n}$ of the attention head for each residual stream then consists of a sum of all of the $\color{red}v_i^{\ell, n}$ vectors to the left of the $\color{red}v_j^{\ell, n}$ vector, attenuated by the attention scores.

\begin{equation}
{\color{teal}o_j^{\ell,n} }= \sum_{i = 1}^j a_j^{(\ell,n,i)}{\color{red}v_i^{\ell,n}}
\end{equation}

All of the individual $\color{teal}o_j^{\ell,n}$ vectors are then concatenated into one $\color{teal}o_j^{\ell:combined}$ for the layer

\begin{equation}
{\color{teal}o_j^{\ell:combined}} = \big\Vert_{n=1}^{n_{head}}{\color{teal}o_j^{\ell,n}}
\end{equation}

The final output of the sublayer is then calculated using the final output feed-forward matrix and bias $W_O^{\ell} \in \mathbb{R}^{d \times d}$ and $b_O^{\ell} \in \mathbb{R}^{d}$,

\begin{equation} 
{\color{teal}O_j^{\ell}} = \text{Norm}((W_O {\color{teal}o_j^{\ell:combined}} + b_O)) 
\end{equation}

\subsubsection{Multi-Layer Perceptron (MLP) Sublayer}
The MLP sublayer consists simply of two feed-forward matrices and biases,
one which expands the input into a vector of size $d_i$, and another which contracts it back down.

\begin{equation}
{\color{Peach}r_{j}^{{\ell-1}:mlpnorm}} = \frac{{\color{magenta}r_j^{\ell-1}} - \mathbb{E}[{\color{magenta}r_j^{\ell-1}}]}{
\sqrt{\text{Var}[{\color{magenta}r_j^{\ell-1}}] + \epsilon}
}*\gamma^\ell_{M} + \beta^\ell_{M}
\end{equation}

\begin{equation}
{\color{gray}m_j^{\ell:i}} = W_1 {\color{Peach}r_{j}^{{\ell-1}:mlpnorm}} + b_1
\end{equation}

\begin{equation}
{\color{brown}m_j^{\ell}} = \text{GeLU}(W_2 {\color{gray}m_j^{\ell:i}} + b_2)
\end{equation}

As a final step, the MLP layer output $\color{brown}m_j^{\ell}$, is then incorporated, alongside the earlier attention output ${\color{teal}O_j^{\ell}}$ into the residual stream  ${\color{magenta}r_j^{\ell}}$ via addition.

\begin{equation}
{\color{magenta}r_j^{\ell}} = {\color{brown}m_j^{\ell}} + {\color{pink}r_j^{\ell - 1}} + {\color{teal}O_j^{\ell}}
\end{equation}

\subsubsection{Dimensions}

There are numerous different dimensions to keep track of when discussing the inner workings of an LLM. The most important of these is the hidden size, $d$, which is the dimension of the majority of the hidden state vectors used by the LLM. There are two other related dimensions, the attention head hidden size, $d_h$, used for hidden states only within an LLM's attention head sublayer, and the MLP intermediate hidden size, $d_i$, which is used only for hidden states inside of the MLP sublayer. In addition, an LLM contains several layers of number $l$, several attention heads per layer of number $n_h$, and a vocabulary of size $|V|$ (see Table \ref{tab:dimensions} for details).

\subsubsection{Self-attention mechanism as symbolic feature copying}

In this section, we provide a view of self-attention in terms of a mechanism that propagates distributed features that can be thought of abstract symbols representing various semantic and grammatical aspects of tokens along the input token sequence all the way to the last token in the input. At that point the symbols accumulated in the vector representation of the last token in the sequence influence the selection of the newly generated token.  Figure~\ref{fig:LLMTopLevel} shows a circuit diagram overview of the entire Pythia-70m model with the prompt "The man I saw yesterday". 
There are two important mathematical-denotative aspects we must emphasize to make sure that crucial details in the following hypothesis narrative are not lost or muddled. First is that, as mentioned before, each token in the prompt creates a "column" of vectors, --- i.e, the residual stream, and that these streams \emph{do not} cross outside of the self-attention heads. Symbols, as decoded by the SAE flow down each of the ``streams'' (vectors).  Furthermore, symbols \emph{also} do not pass intact through transformations, represented by the large grey boxes in Figure~\ref{fig:LLMTopLevel}. The LLM may certainly, over the course of training, learn weights for its transformer matrices such that the matrix's eigenvectors align with the SAE vectors, allowing them to pass through the transformation intact, but this is not what happens by default. This is most relevant in the self-attention layers because the $\color{teal}{o_j^{(\ell,n)}}$ vectors output by the core of the self-attention mechanism must pass through the $W_O,b_O$ transform before making it into the $\color{teal}{O_j^\ell}$ vectors which are added into the residual stream. Any information present in the $\color{teal}{o_j^{(\ell,n)}}$ and the $\color{red}{v_j^{(\ell,n)}}$ vectors which comprise them is interpreted/scrambled/computed on by the $W_O,b_O$ transform. Nevertheless, we expect that substantial quantities of information passes through the $W_O,b_O$ transform ``intact'' in a denotative rather than mathematical sense, for example, that there might be one symbol vector $S_1$ that the model uses to denote grammatical singularity in $\color{teal}{o_j^{(\ell,n)}}$ and  $\color{red}{v_j^{(\ell,n)}}$, another symbol $S_2$ that the model places inside $\color{teal}{O_j^{\ell}}$ vectors meaning the same thing, and a learned $W_O^\ell,b_O^\ell$ that straightforwardly places $S_2$ in the output if $S_1$ is in the input, semantically similar but not mathematically similar in that the elements $S_2$ are quite dissimilar to those of $S_1$.

With a concrete example of English subject-verb-number agreement, we formulate our hypothesis of how Pythia might be executing a fairly simple set of symbolic computations. Our approach in this work has been to start out with the assumption that Pythia-70m has feature vectors denoting concepts relating to grammatical number in its hidden states, induce  which mathematical properties said vectors would have, and then search the model for symbol vectors with the hypothesized properties, and then further examine the discovered vectors for additional properties which were outside of the scope of our initial investigation.

\begin{figure}[H]
  \centering
  \includegraphics[width=0.7\textwidth]{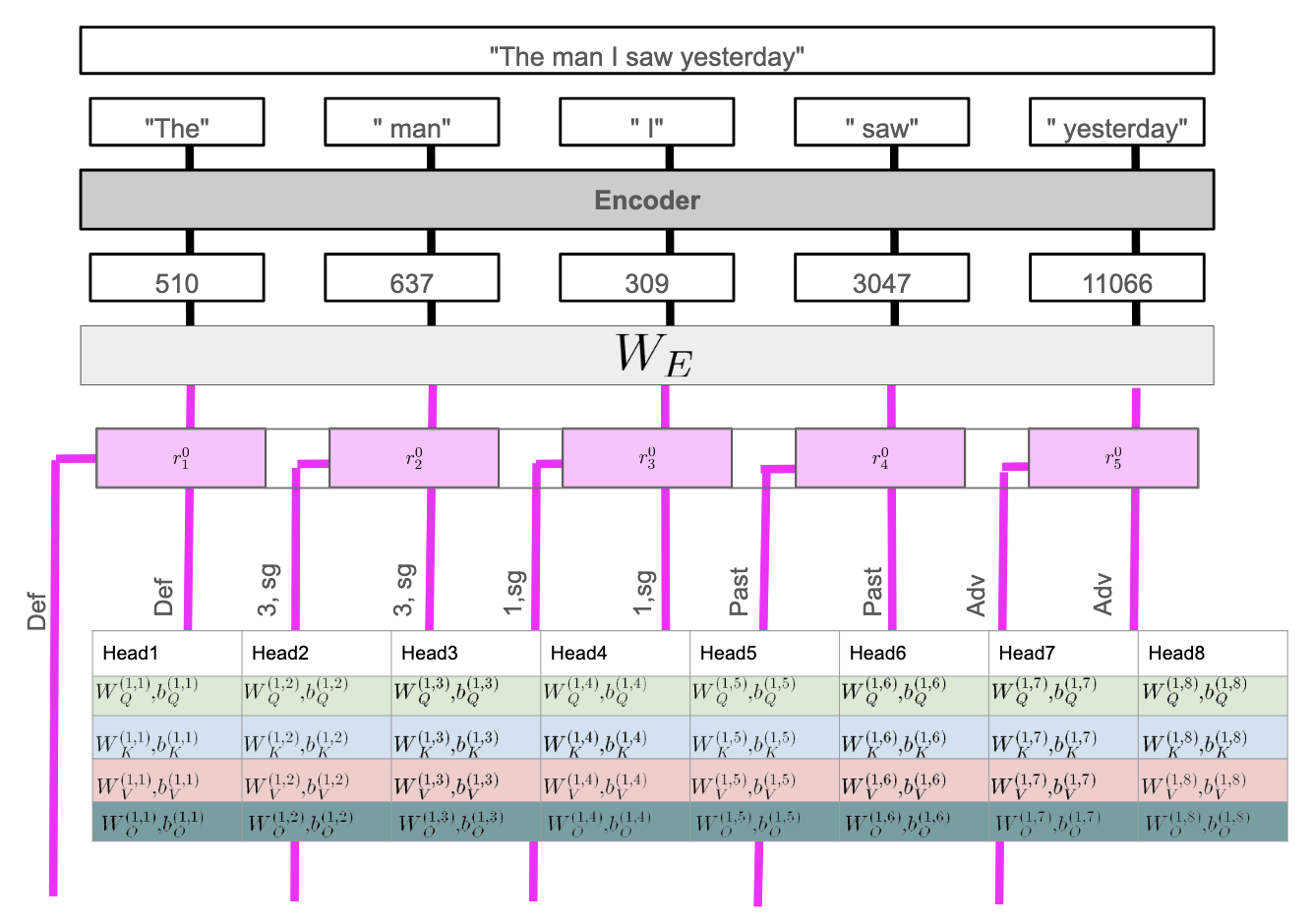}
  \caption{The token `` man'' bearing the \textbf{Sg} features}
  \label{fig:Man3rdPersonSingula}
\end{figure}

\begin{figure}[H]
  \centering
  \includegraphics[width=0.7\textwidth]{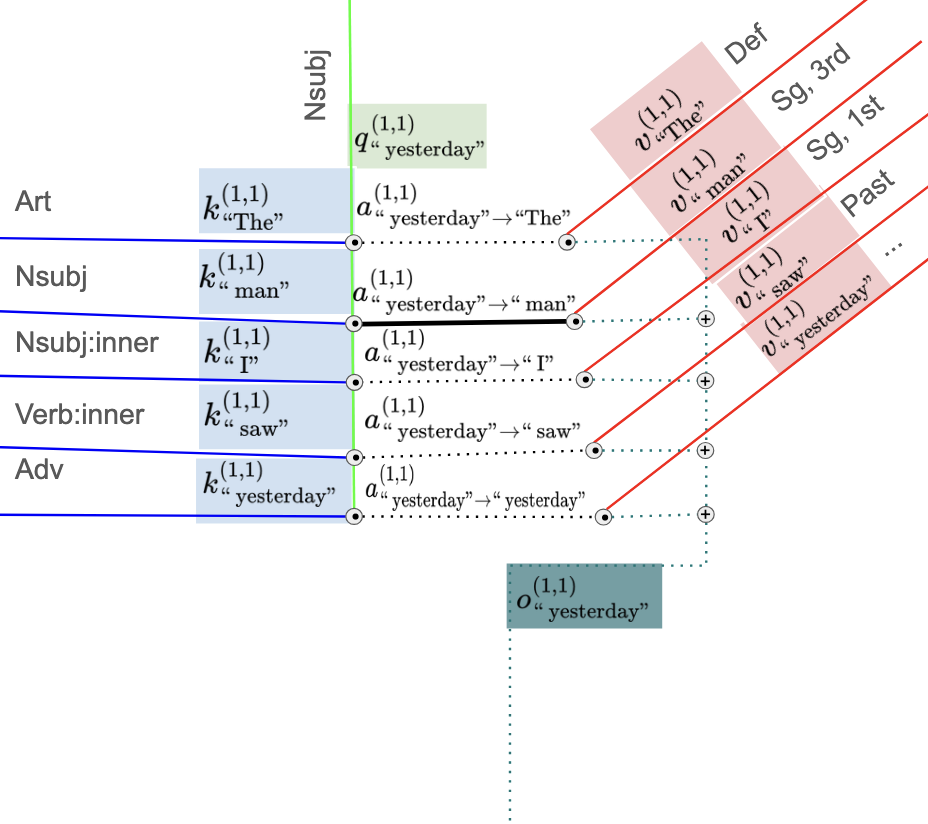}
  \caption{$a_{\text{`` yesterday''} \rightarrow \text{`` man''}^{(1,1)}}$ acquires a high value }
  \label{fig:AttentionGrammar2}
\end{figure}

First, we are going to assume a fairly uncontroversial syntactic mechanism by which the LLM might successfully complete prompts like "The man I saw yesterday" with the token for a singular verb `` is'' instead of a plural verb `` are''.

\textbf{Step 1}. The model will have a series of symbols denoting grammatical singularity, which we collectively call \textbf{Sg} symbols. This begins with a context-free ``singular noun'' symbol, which we will call \textbf{Sg-noun-${\color{magenta}W_E}$} (the last portion indicating which matrix of weights the symbol ultimately originated from) at some point in the residual stream of `` man'' in either the output of an attention head, $o_\text{`` man''}^{(\ell, n)}$, or of the MLP sublayer, $m_\text{`` man''}^{\ell}$  . It is very likely that the singular symbol exists directly in the embedding layer itself, popping out right in the first vector ${\color{magenta}r^0_\text{`` man''}}$, as `` man'' will resolve to a singular noun in most English sentences without context. Another symbol might be used in situations where this is not possible, like on the noun ` deer' in the prompt ``One deer'. Figure~\ref{fig:Man3rdPersonSingula} shows this in circuit representation along with some other plausible grammatical features that might be found on the other tokens.

\textbf{Step 2}. In a later attention sublayer of the model, a certain $\color{red}{v_\text{`` man''}^{\ell,n}}$ will bear yet another semantically synonymous symbol \textbf{Sg-noun-${\color{red}W_V^{(\ell,n)}}$} meaning \textbf{Singular}, calculated from the \textbf{Sg-noun-${\color{magenta}W_E}$} symbol present in the input 
$\color{magenta}{r_\text{`` man''}^\ell}$ (these two Singular symbols will be distinct vectors despite meaning the same thing for reasons discussed earlier). 

\textbf{Step 3} The information hops from the `` man'' residual stream to the final `` yesterday'' residual stream in this step. We expect that in the same attention head as step 2, there will be a pair of key and query vectors, $\color{blue}{k_\text{`` man''}^{\ell,n}}$ and $\color{OliveGreen}{q_\text{`` yesterday''}^{\ell,n}}$, each bearing the same \emph{Nsubj-$({\color{OliveGreen}W_Q^{(n,\ell)}}/{\color{blue}(W_K^{(n,\ell)}})$} symbol which is responsible for copying features from the subject of a verb to the verb itself. This shared symbol raises the dot product of this $\color{blue}k$ $\color{OliveGreen}q$ pair, thus creating a high $a_{`` man''\rightarrow`` yesterday''}^{\ell, n}$ score, opening the ``sluice gate'' between the ` man'' and `` yesterday'' residual streams, so to speak, which will cause the \textbf{Sg-noun-${\color{red}W_V^{(\ell,n)}}$} symbol riding in $\color{red}{v_\text{`` man''}^{\ell,n}}$ to be copied into $\color{teal}{o_\text{`` yesterday''}^{\ell,n}}$. This  is shown below in \ref{fig:AttentionGrammar2}

\textbf{Step 4} The feed forward matrix and bias, after being given the \textbf{Sg-noun-${\color{red}W_V^{(\ell,n)}}$} symbol, in response adds a \textbf{Sg-verb-${\color{teal}W_O^{(\ell,n)}}$} to the output $\color{teal}{O_\text{`` yesterday''}^{\ell}}$ vector, which flows into the $\color{magenta}{r_\text{`` yesterday''}^{\ell}}$ vector, which then flows all the way down to $\color{magenta}{r_\text{`` yesterday''}^{6}}$ and thereafter to the unembedding matrix $W_U$. This symbol would be the ``say a verb'' symbol, to use Anthropic's terminology.

\textbf{Step 5} We then expect that the unembedding vectors for singular verb tokens like `` is'', `` says'', `` does'', etc... will also contain the \textbf{Sg-verb-${\color{teal}W_O^{(\ell,n)}}$} symbol. This symbol being shared between the relevant singular verb token unembedding vectors and $\color{magenta}{r_\text{`` yesterday''}^{6}}$ raises the dot product between them, thus raising the logits, thus elevating the probability at which these tokens are chosen and effectuating the correct verbal inflection choice. This  is shown below in \ref{fig:AttentionGrammar2}

\textbf{Further Complexities} The model may subdivide grammatical singularity or grammatical plurality into several symbols. There may be multiple different attention head hopping mechanisms and asosciated symbols, perhaps dependent on syntactic parsing details. MLP sublayers may aggregate grammatical number representations which are more fragmented in earlier layers into more unified ones to prepare for ``say a verb'' symbol generation. We must have a symbol identification methodology sufficiently broad and thorough to identify symbols involved in these more complex situations.

\section{Methods}

Because some of the methodology we developed and present in this paper, with the exception of how we trained the SAEs, so heavily diverges from the prior efforts of Anthropic's team \cite{ameisen_circuit_nodate} and  Marks et al. \cite{marks_sparse_2025} on which this work is based, we have chosen to include detailed justifications for developing these new methods in addition to their description. While we used Pythia-70m LLM for the development and the initial validation of our approach, we tested the generalizability of the approach on the larger models Gemma-2-2b \cite{team_gemma_2024} and Llama-3.1-8b \cite{grattafiori_llama_2024}, chosen on account of their publicly available SAEs, through Gemma Scope \cite{lieberum_gemma_2024} and Llama Scope \cite{he_llama_2024} respectively.

\subsection{SAE Training}

For Pythia-70m LLM, We trained 12 TopK and Matryoshka SAEs as follows. SAEs were trained on the outputs of each attention and MLP sublayer the ${\color{teal}O_j^\ell}$ and ${\color{brown}m_j^\ell}$ vectors over the fully trained Pythia-70m (checkpoint step 143000), following the methods outlined in \cite{bussmann_learning_2025}. Each of the regular TopK SAEs had a feature space size of 32768 (32 times the hidden state size of 512), whereas the Matryoshka SAEs had a feature space size of 8192. Encoder and decoder biases $b_e$ and $b_d$, respectively, were tied, however, encoder and decoder embeddings $W_e$ and $W_d$ were untied. In addition, we trained a pair of Matryoshka SAEs of feature space size 1024 on the outputs of each individual attention head. One of the pair was trained on the value outputs ($\color{red}{v}$), and the other simultaneously on the key and query vectors ($\color{blue}{k}$ and $\color{ForestGreen}{q}$). We trained the SAEs over the OpenWebText corpus \cite{Gokaslan2019OpenWeb}, with the sentences truncated to a maximum input sequence length of 192 tokens. Matryoshka SAEs were also trained on the embedding layers of the LLMs Gemma-2-2B (16384 features), and Llama-3.1-8B (32768 features), using the same methodology we used for Pythia.

\subsection{Datasets used to examine LLMs linguistic behavior}

To study a syntactically diverse sample of prompts, we used a subset of artificial sentences derived from \cite{marks_sparse_2025}. We used three of their verb conjugation with distractor clause sentence pair collections: a) the prepositional phrase collection (\texttt{nounpp}); b) the matrix clause subject collection (\texttt{rc}), and c) the embedded clause subject collection (\texttt{within\_rc}). We generated ourselves using ChatGPT and manual editing 200 sentence pairs for a new collection of 11-token long \texttt{long\_rc} sentences, which were also all externally-headed relative clauses like \texttt{rc}, but lengthened with more adverbs, adjectives, and other material.

\subsection{Establishing the existence of causal relationships between SAE features and model behavior}

Over the course of our investigations we discovered that there were no computational or analytic shortcuts when it came to investigating the meaning of and relationships between the features in the SAEs we had trained. We instead discovered each of the assumptions listed below to be false, requiring iterative refinement of our previous methods and invention of new ones. By following the narrative below we wish the reader to understand the extent to which our methods differ from those of the prior works which we have cited.

\subsubsection{One verb pair generalizability assumption}

We cannot use a procedure that merely investigates how the model chooses to output one pair of verbs, such as ``is''/`` are'' or ``has''/``have''. This will lead to extensive confounding in our analyses of features the model uses for selecting non-numeric semantic features, including the generation of the present progressive or  perfect tenses in the case of the preceding pairs, and the numeric agreement features we are actually interested in. Thus, $B$ must take many verb pairs into account.

\subsubsection{One sentence pair generalizability assumption}

For similar reasons, detailed investigation of the model's number agreement behavior on a single sentence pair, like ``The child plays''/``The children play'' may confound abstract number agreement phenomena with semantic and syntactic phenomena specific to the sentence in question. As discussed later, we indeed found this to be the case, feature we:108 seemed to be indexing specifically human plural nouns, e.g., ``children'', rather than plural nouns in general, and feature 4:attn:429 is only used by the model to generate plural verb tokens when there are tokens intervening between the subject and the verb.

\subsubsection{Gradients of verb conjugation as substitute for activation patching (AP) assumption}

Calculating the gradients of a model training step designed to invert verb conjugation is far cheaper computationally than the effects of full-sized AP's. Due to the nonlinear nature of LLMs, it is not true that the effect of an infinitesimal introduction of a small AP at strength $b$ introducing a feature $f$ on a prompt $P$ has a predictable relationship with what would happen if $b$ were set to a realistically large value $\Delta$, i.e.,  $\frac{\partial \space \text{effect}(P|P_h = P_h + bf)}{\partial b} >0 \notimplies \text{effect}(P|P_h = P_h + \Delta f) > 0 $. It is indeed the case that, if the gradient of an AP with the effect ``promotes singular verb conjugation'' is positive, then the full-sized effect of such an AP \textit{is somewhat likely to} promote singular verb conjugation. If this were not the case, gradient descent training of LLMs would accomplish little; however, we still found that this relationship was weaker than one might expect.  

The weakness of this relationship was most prominently visible in our experiments aimed at identifying singular and plural noun-number features. We experimented with applying AP's on the \texttt{nounpp} and \texttt{rc} datasets searching for AP's which would flip the number of generated verb tokens based on searching for the top 60 candidate AP's with the strongest gradient effect sizes. For example, the sentence ``The AP(man) in the park (is/are)'', readily yielded context-free 0:mlp features which, when set on singular nouns, caused plural verb conjugation, these being \texttt{0:mlp:47}, \texttt{0:mlp:134}, \texttt{0:mlp:829} (See the results on the marking of grammatical number on noun tokens for further discussion). However, the same experiment, when performed on the \texttt{within\_rc} dataset, i.e., ``The man that the the AP(dog) (is/are)'', yielded only \texttt{0:mlp:134}. Despite this, through another round of analysis, we found that the other 2 aforementioned plural noun features did in fact have a strong full-sized AP pluralizing verb conjugation effect when applied to embedded noun subject tokens in the \texttt{within\_rc} dataset anyway. From this we learned that the first gradient search technique may miss important AP's and also yield irrelevant ones.

\subsubsection{Feature effect relevance assumption}

We discovered a feature, \texttt{4:mlp:961}, which passed all three preceding assumption hurdles, strongly and consistently promoting plural verb conjugation as measured by our earlier-outlined metric when set on the final token of each of the prompts of the \texttt{rc} dataset.

We then investigated the ordinary distribution of the strength of this feature on the un-AP'ed final tokens of each sentence in the aforementioned dataset and the two others, \texttt{within\_rc} and \texttt{nounpp}, as affected by the grammatical number of the matrix and embedded noun subjects in said sentence.

\begin{figure}[H]
  \centering
  \includegraphics[width=\textwidth]{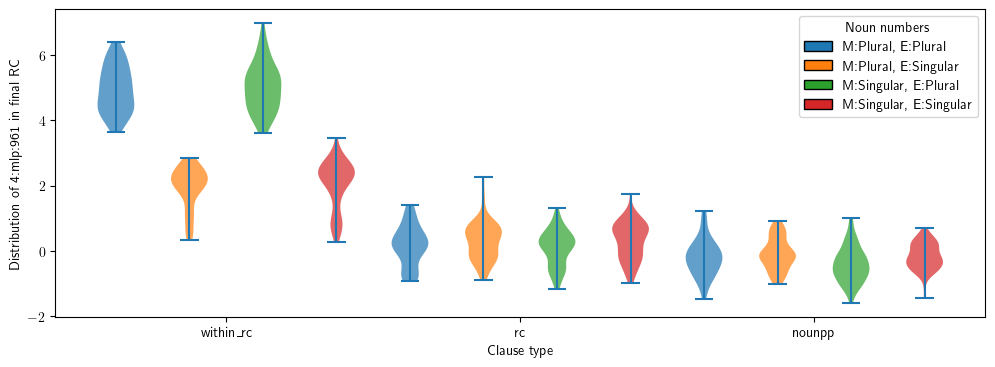}
  \caption{4:mlp:961 tracks embedded clause noun number. M = Matrix noun, E = Embedded noun}
  \label{fig:mlp_4_961}
\end{figure}

This revealed, to our surprise, that \texttt{4:mlp:961} is not used at all by Pythia-70m for the long-distance number agreement in object-headed relative clauses. We used the datasets \texttt{rc} and \texttt{nounpp} to investigate this type of agreement and preliminary experiments revealed that 4:mlp:429 seems to be used by the model for conjugating over subject-headed relative clauses. However, we did not do a full investigation. We learned from this preliminary experiment that we needed to check that a candidate feature with a hypothesized property, such as ``causes the generation of a plural verb token'', exhibits a \textit{bimodal} distribution in ordinary model operation, being set to one narrow range of values in appropriate contexts for the hypothesized property and another non-overlapping range of values in inappropriate contexts. 

\subsection{Quantifying the existence of a causal relationship from one feature or a token substitution to either another feature or a model's behavior}

To avoid getting lost in this forest of caveats and considerations, we had to be very purposeful and precise in formalizing statements such as ``Feature 4:attn:429 \textit{means} X.'', or ``Feature 0:mlp:134 \textit{means} Y.'' Using the Pearl's causal framework discussed in the background section, we devised several metrics and a procedure by which a set of SAE features involved in a given LLM behavior could both be identified and their relationships to each other quantified. The application of these metrics and procedures thus allows us to make precise formal statements such as ``SAE feature 0:mlp:47 is distinctively marked on animate plural noun tokens and is partially responsible for causing plural verb agreement in subsequent agreeing verbs''. This statement refers to a specific \textit{behavior} in which Pythia-70m exhibits a tendency to output singular verb tokens when 0:mlp:47 is present on a sentence's noun subject token, which is reduced when this feature is removed, and is promoted when this feature is added onto noun subject tokens it is not normally on.

First, since we suspected a priori that there would be significant variety in model behavior depending on the syntactic structure of the prompt, we divided our prompts for investigation into separate datasets, as discussed previously. We refer to a nonspecific member of these datasets (\texttt{rc},\texttt{long\_rc},\texttt{within\_rc},\texttt{nounpp}) with the set variable $S$ in the subsequent equations. Each set $S$ consists of a singular subject corpus $S_s$ and a plural subject corpus $S_p$. The sentences in these corpora were paired into $s_s,s_p$ pairs such as ``The \textit{man} that the dog bites...'' and ``The \textit{men} that the dog bites...'', differing only in the number of the subject in order to prevent our process from being distracted from non-numerical semantic differences in the underlying nouns and verbs. We generated the sentences in these corpora such that all had the same number of tokens, which enabled us to calculate much more quickly in parallel using batching, at the cost of preventing us from investigating mechanisms held in common across a variety of sentence lengths.

We can also formalize a hypothesis $H$ about a feature $F$, such that ``Feature F being present at a high value $b$ on the final token promotes the generation of plural verb forms and is itself caused by the matrix subject being token plural.'' and examine the following scenarios:

\begin{enumerate}
\item Establishing the endogenous variables in $V$ 
\begin{itemize}
    \item In theory, every last feature $F$ in our sparse autoencoders should be part of our causal model. Investigating them all  is computationally infeasible.
    Therefore, we first created a basic filter for identifying features which were at least likely to be involved in verb conjugation
    due to having non-overlapping distributions between  a group of singular prompts $s_s$ like ``The dog that the woman sees'' and a group of plural prompts $s_p$ like
    ``The dogs that the woman sees...''. Figure~\ref{fig:context_free_number_features} shows a visualization of what these filtering calculations look like. We collected large quantities of feature activation strengths for a variety of singular (green) and plural (blue) nouns and looked for features whose activation value distributions did not overlap between the two sets, as measured by their Cliff's delta scores (see a more in-depth discussion of this score in section 3.7). When all plural activations were higher than all singular activations, the feature's Cliff's delta score is 1; -1 if all plural activations were less than the singular ones; and an intermediate score if there was overlap in the activation distributions.

\end{itemize} 
\item Quantifying verb conjugation with the behavior function $B$
\begin{itemize}
\item Rigorous analysis of LLM behavior requires us to measure its behavior with scalar. Therefore, we needed to devise a function over the token probability distribution output by the LLM which quantifies the model's preference to produce singular instead of plural verbs in response to a given prompt, which we termed $B$, or the ``singular conjugation preference metric''. Developing this function proved to be one of the most challenging aspects of this study.
With $B$, we were able to further idenfify features $F$ for $V$ that actually matter for model verb conjugation. $B$ is much more expensive to calculate than the Cliff's delta so we used it as a second filtration step. Figure \ref{fig:0MLP134B} shows what happens to the $B$ score in one of our datasets when we performed the operation $B(s_s| \texttt{do}(\texttt{0:mlp:134} = \texttt{0:mlp:134}(s_p))$ and vice versa (0:mlp:134 seems to be a plural marking feature on nouns), in an attempt to ``trick'' Pythia-70m into thinking that nouns with singular subjects had plural subjects and vice versa, the quantity of trickery measured by how much singular verb conjugation $B$ was altered. We found, by manipulating this feature into counterfactual values, that it had a strong effect on verb conjugation in both cases but that it was easier  to ``singularize'' prompts by lowering the values of 0:mlp:134 than it was to ``pluralize'' prompts by raising its value.

\end{itemize}
\item Finding important edges for the causal model $E$
\begin{itemize}
    \item Every cause feature $F_c$ appearing in a model's hidden state vectors has some effect on each effect feature $F_e$ occurring in both a subsequent layer and either in the same or subsequent token's residual streams. The vast majority of these $F_c \rightarrow F_e$ relationships are weak and unimportant and so we want a means of identifying the ones that do matter.
    The work in \ref{fig:0MLP134-5Attn966} shows an example of such an edge between $F_c$ = 0:mlp:134 (which appears to be a plural marking feature on subject nouns, such as on the token `` dogs'' in ``the \textit{dogs} that the women sees...'') and $F_e$ = 5:attn:966, a feature which is involved in selecting plural verbs, showing up on the final token ``sees' in ``the dogs that the woman \textit{sees}...''.
    The distributions of $F_e(s_s |do(F_c=F_c(s_p)$ (an attempt to ``force'' Pythia-70m to imagine the token ``dogs'' to have the singular number) and of $F_e(s_p |do(F_c=F_c(s_s)$ (an attempt to ``force'' the model to instead imagine that ``dog'' has the plural number doing the inverse) are dramatically different. We are interested in identifying only those $F_c \rightarrow F_e$ relationships that are ``important'' enough to show these dramatic differences.

\end{itemize} 

\end{enumerate}

This hypothesis framework is a purely behaviorist account of what it means for an SAE feature in an LLM to ``mean'' something, with prompts as stimuli and generated tokens as responses. LLM-generated language cannot ``denote'' in the linguistic sense of the word. Lacking bodies or sense organs, LLMs cannot refer to people, places, or things that they have encountered in their environment. LLM language is pure anaphora, all discourse entities referring back directly or drawn in by implication to either earlier snippets of language introduced by the user or the LLM itself (an LLM makes no robust architectural distinction between the language which it itself has said and the language typed by the user), or to generic world knowledge, real and fictional, implicitly encoded in the LLM's learned weights. The world knowledge itself only manifests in a behavioral form, as certain statements generated in response to certain prompts, not as a human-legible database carefully maintained as to purge contradictions, hence, LLM's notorious hallucinatory tendency to answer the same question in the positive when it is phrased one way and in the negative should it be rephrased in some other ever-so-slightly different way. 

\subsection{The meaning of ``meaningless" SAE features found in an untrained LLM} With this framework, we can then formalize what it means for the features found by an SAE trained on an untrained LLM such as Pythia-70b step 0 checkpoint to be ``meaningless'', as this untrained model still responds to prompts, albeit in an unhelpful way. It is true that certain sentences $s$ will cause certain features $F$ to appear, causing certain tokens $T$ to be generated, however, in an untrained LLM, there is no human-discernible rhyme or reason to the relationships between $s$ and $s\prime$, or $T$ and $T\prime$. By saying that untrained LLM's generated language and its features are ``meaningless' we mean to say that the language it generates is devoid of usefulness and patterns.

\subsection{Quantifying LLM Verb Conjugation Behavior with $B$}

Developing our singular conjugation preference metric $B$ proved to be the most challenging primarily due to the LLM's tokenizer. 

From a human linguistic perspective, the rules of English number agreement are relatively simple. To make a verb singular, add a suffix to the verb which is spelled``s'' or ``es'' and pronounced /s/, /z/, or /\textsc{i}z/ following a fairly simple set of rules to make the correct choice of suffix variant. It is not so simple with an LLM. Although the token ``s'' is in Pythia as token \#84, this token is rarely produced, especially for common verbs. Instead, a fully conjugated portmanteau token, or a sequence of word parts, is produced, depending on the verb in question. For example, `` go'' is token\#564, whereas `` goes'' is \#4566. `` grab'' is \#10013 and `` grabs'' is \#42562. For hundreds of common verbs in English, the LLM is forced to memorize these arbitrary relationships, similarly to how English speakers memorize irregular verb forms such as ``is/are''. For less common verbs, the situation is even more complicated. The tokenizer does not divide uncommon words in ways which respect morphology. For example, `` uphold'' is the single token\#39954, whereas `` upholds'' is broken up into \#598, \#15532 (`` up'', ``holds''). Furthermore, alternations in capitalization and spacing also change the tokenization.

\subsubsection{Defining a set of verb form pairs not subject to irregular tokenization}

From the outset, we know $B$ has to take into account at the very least hundreds of singular and plural verb token probabilities to make a decent approximation of the abstract notion of ``singular conjugation preference''. We used the following process to gather the verb form tokens which $B$ would make use of. 
First, we immediately excluded from consideration verbs which get split into multiple tokens in one or the other of their singular or plural inflected forms, as calculating $P(\text{verb})$ in parallel over a large set of verbs is exceedingly expensive whenever the verbs consist of more than one token. 

Second, to avoid the confusion of capitalization and spacing token variants, we only included verb tokens which were all lowercase and had a leading space, by far the most common variety. This is because, while Pythia has the full set of 6 variant tokens for the verb form ``is'' ( `` is'', ``is'', `` Is'', ``Is'', `` IS'', and ``IS''), almost no other verb forms have the full set. For example, `` say'' is in the model but ``SAY'' is fragmented into ``SA'' ``Y''. 

Third, we excluded from consideration syntactically ambiguous tokens that were likely to be used as nouns instead of verbs, a frequent phenomenon in English. Consider the sentences ``The city houses ten thousand people.''  and ``The city houses are smaller than the country houses.''. In one, ``houses'' is a singular verb, in the other, a plural noun. Since we can't tell from a single token generation step whether Pythia ``meant'' to generate a noun or a verb, we chose to exclude the most highly ambiguous in isolation (without the right-hand context available) noun-verbs from consideration. We determined which tokens were usually nouns, or usually verbs, by having Spacy's \texttt{en\_core\_web\_sm} parser \cite{explosion_spacy_english} lemmatize and tag the first 100,000 sentences of the corpus we used to train the SAEs, collecting those lemmas which were parsed as being used as verbs in context in more than 75\% of instances (or conversely as nouns more than 75\% of the time), to filter out syntactically ambiguous tokens. We found 367 suitable monotokenal verb pairs for $\mathbb{V}$ and 1700 noun pairs for $\mathbb{N}$.

With these criteria in mind we gathered a large sample of verb form pairs $\mathbb{V}$ (and also noun form pairs $\mathbb{N}$ for later usage) for $B$ to take into consideration. Each sample consists of a set of pairs $(v_{s},v_{p})$ or $(n_{s},n_{p})$ such as `` is'', `` are'' or `` child'', `` children''. We identified 367 verbs for $\mathbb{V}$, and 1700 nouns for $\mathbb{N}$ such that both the singular and plural forms existed as single lowercase leading space tokens in the LLM's vocabulary. 

\begin{figure}[!htb]
\makebox[\textwidth][c]{%
  \includegraphics[width=1\textwidth, keepaspectratio]{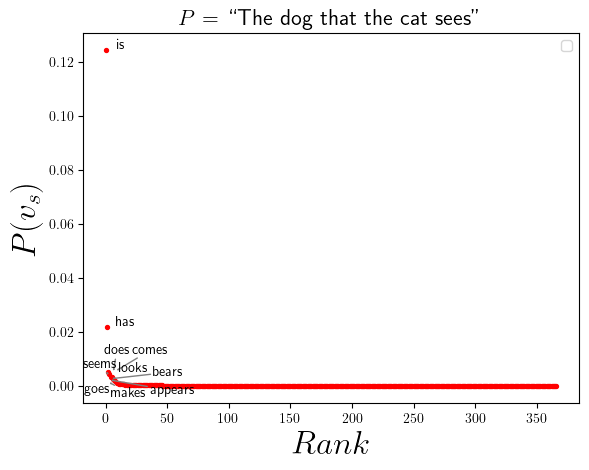}
  }
  \caption{Illustration of the Zipfian nature of the verb token generation probabilities distribution}
  \label{fig:zipfian}
\end{figure}

\begin{figure}[!htb]
\makebox[\textwidth][c]{%
  \includegraphics[width=1\textwidth, keepaspectratio]{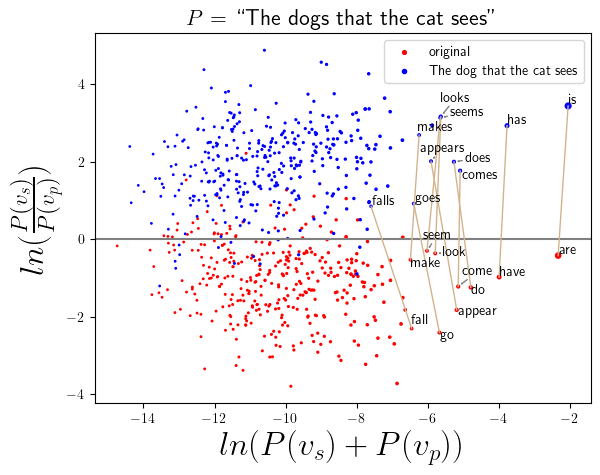}
  }
  \caption{Illustration of the structure revealed by using log probabilities}
  \label{fig:logprobzipfian}
\end{figure}

Fourth, having identified the $\mathbb{V}$ tokens that we wanted $B$ to consider, we then had to design a function to map a list of token probabilities to a single number. The token probabilities output by Pythia exponentially decay by rank, as shown in figure \ref{fig:zipfian}. The decay follows a Zipfian distribution, which is unsurprising, given that the Zipfian distribution as shown in figure \ref{fig:logprobzipfian} was originally discovered by examining English word frequencies. Given these consideration, we chose to analyze these probabilities in log-space. If we did not analyze token probabilities to log-space, $B$ would be dominated by the model's preferences to generate the following 3 auxiliary verb form pairs: ``is/are'', ``has/have'', and ``do/does'', ignoring almost all other tokens. Since auxiliary verbs have unique grammar and semantics in addition to number marking, we did not want $B$ to be severely affected by these considerations. Consequently, we chose $ln(\frac{P(\text{`` is''})}{P(\text{`` are''})})$, $ln(\frac{P(\text{`` has''})}{P(\text{`` have''})})$, $ln(\frac{P(\text{`` does''})}{P(\text{`` do''})})$, ... $ln(\frac{P(v_s)}{P(v_p)})$ (which is equal to $ln(P(v_s)) - ln(P(v_p))$) as our measure of how much the model ``wanted'' to say a singular verb form as opposed to its plural counterpart. Figure \ref{fig:logprobzipfian}, which shows this metric over $\mathbb{V}$ measured on a singular subject/ plural subject pair of sentences, shows that this value for each verb form pair (one dot per pair) is stable (i.e., does not explode to infinity or collapse to 0).


Fifth, we needed to prevent statistical noise from ultra-low probability tokens from corrupting the output of $B$. Pythia-70m does not always assign higher probabilities to grammatically correct verb tokens instead of grammatically incorrect ones, particularly when the probability of either is very low. For example, in response to the prompt ``The dogs that the cat sees...'', Pythia-70m ``prefers" to continue with the incorrect singular verb ``The dogs that the cat sees shoots'' (with probability 0.00007) than the correct plural verb ``The dogs that the cat sees shoot'' (with probability 0.00005). Every point at which Figure \ref{fig:logprobzipfian} has a plural verb form (red dot) above the 0 line, or a singular verb form (blue dot) below the line represents a situation where Pythia prefers the grammatically incorrect form, and thus would assign a higher perplexity to the grammatically correct form than the incorrect. This is common in the low probability verbs but does not occur in the high probability ones, which is why Pythia does not make these sorts of grammatical errors unless its temperature in generation is set quite high. This demonstrates that grammatically correct but semantically unusual sentences confuse Pythia and thus our metric to quantify the model's verb conjugation behavior should downweight them to avoid being overwhelmed by statistical noise.

Through an initial round of unstructured exploratory analysis of the effects of various activation patching (AP'ing) procedures on the model's probabilities of producing verb tokens from the above subset, we empirically determined a satisfying means of mathematically isolating probability distribution changes which corresponded to verb number agreement changes while penalizing extraneous alterations to model behavior. 

With these considerations in mind we settled on the following metric for quantifying an AP's effect $e$ on causing the model to output the singular form, which we term as either as ``the singular conjugation preference between the sentences $s_s =$`The dog that the cat sees.' and $s_p =$`The dogs that the cat sees.' or more briefly $\text{Prf}_s(s_s, s_p)$. In our empirical measurements, a typical value for $\text{Prf}_s(s_s, s_p)$ is 2-3, when the singular/plural sentence pair is ``easy'' for the LLM to conjugate, whereas this number is closer to 0 when phenomena such as number distraction are providing challenges to the model. 

The raw effect for a given verb pair simply measures the log of the ratio of how strongly the model prefers to conjugate in the singular instead of the plural for a given pair after patching.

\begin{equation} \label{Eq:raweffect}
\text{r-effect}(v, \text{patch}) = \text{ln}(\frac{P^{patched}(v_s)}{P^{patched}(v_p)})
\end{equation}

\begin{figure}[!htb]
\makebox[\textwidth][c]{%
  \includegraphics[width=1\textwidth, keepaspectratio]{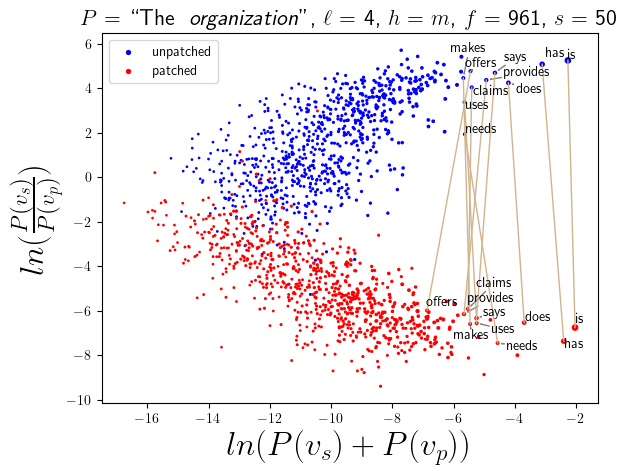}
  }
  \caption{Effect of Layer 4 MLP Matryoshka SAE feature number 961 on Pythia-70m verb conjugation behavior when applied to token `` organization'' at a strength of 50}
  \label{fig:LogFeatureEffect}
\end{figure}

Figure~\ref{fig:LogFeatureEffect} provides a visual intuition and motivation for our singular conjugation preference formula in Eq.~\ref{Eq:raweffect}. The x-axis is the log probability of the model generating any verb form (singular or plural), and the y-axis is the log of the ratio of the model's probability of generating the singular verb form to the probability of generating the plural verb form. 

\subsubsection{Refinements of $B$ to account for verb distributional characteristics}

We also noticed during preliminary exploration that those verb lemmas which the model generally ``wants'' to or ``prefers" to generate (e.g., ``makes'' or ``says'' or ``is'' to continue the prompt ``The organization ...'') are also more likely to be conjugated correctly. This behavior is reflected in the upward diagonal distribution of the blue dots, each representing a single verb pair. From this we learned that our conjugation metric should still weight frequent verbs somewhat more heavily than infrequent ones, whose conjugation behavior is more noisy. Finally, the model's preferences of \textit{which} verb to produce should change only which of the verb's \textit{forms} is chosen. Based on these observations we introduced three attenuating mechanisms: pair penalty, nonverb penalty and weight penalty.

The pair penalty, formalized in Eq. ~\ref{Eq:pairpenalty}, attenuates the AP patch by either increasing or decreasing in total the probability of outputting either of the verb forms. It is 1 if there is no change and less than 1 otherwise.

\begin{equation} \label{Eq:pairpenalty}
\text{p-penalty}(v, \text{patch}) = e^{ln^{\text{abs}(\frac{
ln(P^{\text{patched}}(v_s) + P^{\text{patched}}(v_p))
}{
ln(P^{\text{unpatched}}(v_s) + P^{\text{unpatched}}(v_p))
})}}
\end{equation}

The nonverb penalty in Eq.~\ref{Eq:nonverbpenalty} performs the same calculation as in Eq.~\ref{Eq:pairpenalty} but summed over all nonverb tokens $a$, penalizing patches which increase or decrease the probability of outputting any of the tested one token verbs.

\begin{equation} \label{Eq:nonverbpenalty}
\text{nv-penalty}(\text{patch}) = e^{ln^{\text{abs}(\frac{
ln(\sum_{a \in \text{nonverbs}}P^{\text{patched}}(a))
}{
ln(\sum_{a \in \text{nonverbs}}P^{\text{unpatched}}(a))
})}}
\end{equation}

The weight penalty in Eq.~\ref{Eq:weightpenalty} downweights verbs which the model is unlikely to ever output for the given prompt.

\begin{equation} \label{Eq:weightpenalty}
\text{w-penalty}(v) = \frac{-1}{\text{ln}(P^{\text{unpatched}}(v_s) + P^{\text{unpatched}}(v_p))}
\end{equation}

The total score is then evaluated, per prompt, on each verb pair in our set as follows in Eq.~\ref{Eq:totalscore}

\begin{equation} \label{Eq:totalscore}
\begin{split}
\text{score}(\text{patch}) = {} & \sum_{v \in \text{verbs}} \Bigl[ \text{r-effect}(v, \text{patch})
  \cdot \text{p-penalty}(v, \text{patch}) \cdot \text{w-penalty}(v) \Bigr] \\
  & \cdot \text{nv-penalty}(\text{patch})
\end{split}
\end{equation}

\subsection{Identifying candidate important endogenous variables with Cliff's delta}

\begin{figure}[H]
  \centering
  \includegraphics[width=\textwidth]{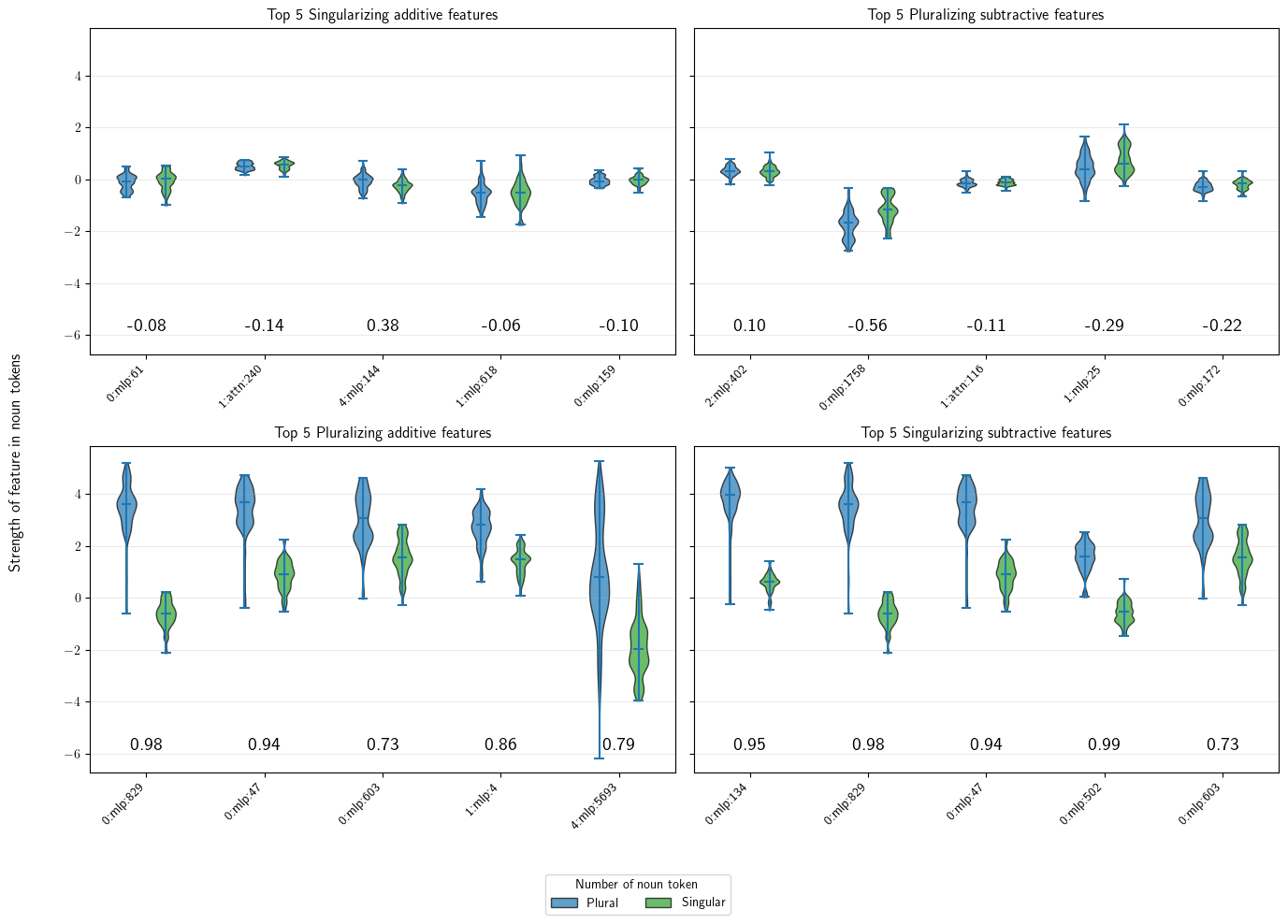}
  \caption{Asymmetry in plurality vs. singularity marking on noun tokens. Cliff's delta scores (-1 to 1 range) are shown beneath each pair of ranges (blue=plural; green=singular) in the strength of the features on the x-axis.}
  \label{fig:context_free_number_features}
\end{figure}

Features which are likely to be important endogenous variables can be quantified by measuring how much overlap there is in the distribution of the values $b \in \mathbb{B}$ of a feature $F$ as it appears. Let $\mathbb{B}_s={F(s):s_s\in S_s}$, hence notated $\mathbb{B}_s=F(S_s)$. Let $\mathbb{B}_p=F(S_p)$. We want to know ``Does $F$ distinguish $S_s$ and $S_p$?''? If $\mathbb{B}_s$ and $\mathbb{B}_p$ do not overlap much, or at all, the preceding statement is more supported. This separation can be measured with the fewest possible assumptions about the shape of the underlying distributions of $\mathbb{B}_s$ and $\mathbb{B}_p$ using Cliff's delta, which simply measures the likelihood that any random $b_s \in \mathbb{B}_s$ will be greater than any random $b_p \in \mathbb{B}_p$. We generated Figure~\ref{fig:context_free_number_features} while looking for features that distinguished singularity and plurality on nouns during our methodological development phase ($S$ was a collection of singular nouns prepended by the token ``The'' and $S+p$ plural nouns in the same setup). This figure illustrates the intuition of Cliff's delta, whose value for each $\mathbb{B}_s,\mathbb{B}_p$ pair (represented as a blue and green violin pairs) is printed below them. When the violins do not overlap much, this value is approaches to 1 or -1 depending on which is higher, and approaches 0 if the two distributions overlap heavily. 

First, from the hypothesis $H$, such as, ``Feature 4:attn:429 being present on the final token promotes the generation of plural verb forms and is itself caused by the matrix subject being a plural token.'', devise a set of test sentences $S$ such that each sentence $s \in S$ is of the correct shape to test the hypothesis. For this $H$, the sentences from the \texttt{rc} (``The man that the dog sees...'') and \texttt{nounpp} datasets (``The man in the park...'') are the correct shape, and thus $S = \texttt{rc} \cup\texttt{nounpp}$.

We used Cliff's delta to assess the separation of feature distributions. We considered a feature to be a potential endogenous variable worthy of further investigation if the Cliff's delta of between the $s_s$ and $s_p$ prompts in a dataset $S$ was greater than 0.7 or less than -0.7.

\subsection{Quantifying Relationships between Endogenous Variables}

There are 2 types of edges between the endogenous variables in the causal graphs we use to describe the semantics of SAE features: 1)Cause SAE Feature - Effect SAE Feature ($F_c \rightarrow F_e$) and 2) Cause SAE Feature - Verb Conjugation ($F_c \rightarrow B$)

\begin{figure}[!htbp]
  \centering
  \includegraphics[width=1\textwidth]{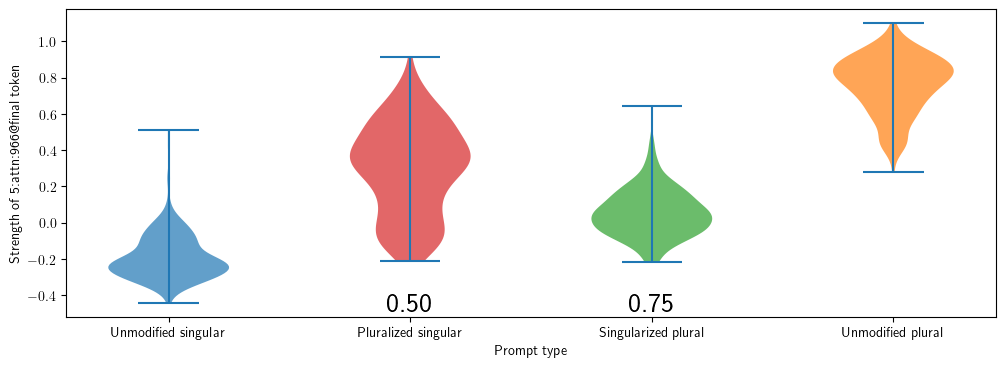}
  \caption{The strength of the $F_c$ = 0:mlp:134@noun subject $\rightarrow F_e$ = 5:attn:966@final token edge over $S=\texttt{rc}$. The numbers under the pluralized singular and the singularized plural violins are Cliff's delta values.}
  \label{fig:0MLP134-5Attn966}
\end{figure}

The Feature-Feature edges are easiest to explain. Consider an $s_s, s_p$ pair: ``The dog in the park...'' ``The dogs in the park''. We are testing if cause feature $F_c$ 0:mlp:134 on token 1 (``dog/dogs'') causes changes in the values of effect feature $F_e$ 5:attn:966 on token 4 (the final token). We collect the ordinary singular and plural values of these features, $F_c(s_s), F_c(s_p), F_e(s_s), F_e(s_p)$. The latter two effect values we term $\text{score}_s(s_s)$ and $\text{score}_p(s_p)$. We then collect the ``singularized plural'', $\text{score}_{p\rightarrow s}(s_s, s_p)$, $F_e(s_p|do(F_c:=F_c(s_s)))$ and ``pluralized singular'', $\text{score}_{s\rightarrow p}(s_s, s_p)$, $F_e(s_s|do(F_c:=F_c(s_p)))$ values of $F_e$ through activation patching. The ``singularization strength'' of the $F_c \rightarrow F_e$ relation, $\text{strength}_s(F_c \rightarrow F_e,s_s, s_p)$ is  $\frac{\text{score}_{p\rightarrow s}(s_s, s_p) - \text{score}_p(s_s, s_p)}{\text{score}_s(s_s) - \text{score}_p(s_s, s_p)}$. This is the fraction of $F_e$'s value in the singular sentence explained by swapping around the value of $F_c$ from its singular to its plural value in the plural sentence. The corresponding ``pluralization strength'', $\text{strength}_p(F_c \rightarrow F_e,s_s, s_p)$, is $\frac{\text{score}_{s\rightarrow p}(s_s, s_p) - \text{score}_s(s_s, s_p)}{\text{score}_p(s_s, s_p) - \text{score}_s(s_s, s_p)}$. We  extend this definition to the entire testing corpus $S$,  $\text{strength}_p(F_c \rightarrow F_e,S)$ by averaging the strengths obtained over all pairs. This score is almost always between 0 and 1. Figure~\ref{fig:0MLP134-5Attn966} shows the distributions of strengths of the $F_e$ effect feature 5:attn:966 appearing in the last token in all of the $s_s$ or $s_p$ sentences of $S=\texttt{rc}$, with scores for each of the strengths printed below. The singularized plural sentences resembled the singular sentences more than the pluralized singular sentences resembled the plural sentences, hence the higher average  $\text{strength}_s(\text{0:mlp:134@noun subject} \rightarrow \text{5:attn:966@final token},s_s, s_p)$ score of 0.75.
The calculations for $F_c \rightarrow B$ merely substitute $B$ for $F_e$ and are otherwise identical. The $F_c \rightarrow B$ edges are ``fuzzy'', they do not distinguish if $F_c$ causes the conjugation behavior directly or if it instead intermediated with other features, or some mixture of the two. The existence of an $F_c \rightarrow B$  edge merely demonstrates that $F_c$ is involved some way or another in verb conjugation. We indeed ended up finding some ``mystery'' features that had an $F_c \rightarrow B$ edge but were wholly disconnected from the $F_c \rightarrow F_e$ graph . Because of how expensive these edges are to calculate, we used the existence of sufficiently strong $F_c \rightarrow B$ edges as a filter condition to determine which $F_c \rightarrow F_e$ edges were actually worth calculating.

\begin{figure}[!htbp]
  \centering
  \includegraphics[width=1\textwidth]{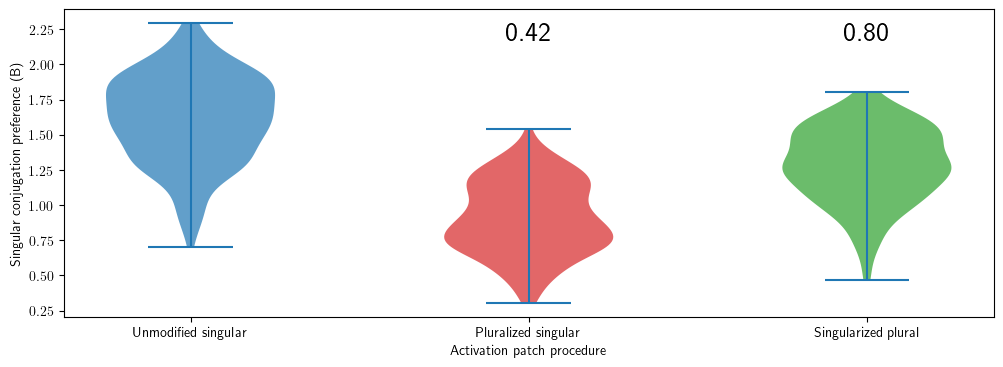}
  \caption{The strength of the $F_c$ = 0:mlp:134@noun subject $\rightarrow B$ edge over $S=\texttt{rc}$. The numbers above the pluralized singular and the singularized plural violins are Cliff's delta values. }
  \label{fig:0MLP134B}
\end{figure}

$\text{strength}_p(F_c \rightarrow B,S)$, the effect of the causal feature on verb conjugation behavior, can be simply obtained by replacing all $F_e$ values with the singular conjugation preference $B$ values caused by the activation patch. This score is almost always between 0 and 1.  The violin plot of $B$ scores obtained over the sentence pairs in \texttt{rc} under varying conditions shown in Figure~\ref{fig:0MLP134B} corresponds to a strong $\text{strength}_p(F_c \rightarrow B,S)$ edge.

Any $W_E$ features which both show a significant Cliff's delta effect size between the singular and plural prompt nouns and have a strong $F_C \rightarrow B$ edge to the output form the direct connection between noun number in the token vocabulary and number marking, as nothing is upstream of $W_E$. It should also be noted that the 0:mlp features are peculiarly connected to the $W_E$ features. The 0:mlp sublayer operates before any attention mechanism has taken effect; therefore, the outputs of the 0:mlp layer are actually pure functions of the row vectors of $W_E$. Any $W_U$ features that have a strong $F_C \rightarrow B$ edge are directly connected to verb form choice, as nothing is downstream of them.

\subsection{Procedures for automatically generating SAE Feature verb conjugation diagrams}

Values for the above scores and strengths can be calculated for all of the features in our SAEs, but to get interpretable circuit diagrams, we need to aggressively prune the features. In addition, because many of these computations are quite expensive, we need to screen for most promising candidate features before proceeding with the full set of calculations as described in the following section.

\subsubsection{Screening}

Features with a Cliff's delta between $S_s$ and $S_p$ less than 0.7 were excluded from further consideration. In addition, any feature $F$ such that the absolute value of  $\frac{\partial B(s)}{\partial h_i} \cdot SAE(H)_F$ was not in the top or bottom 100 for each index $i$ in the prompt was excluded. This heuristic was used as a computationally cheap means of determining if the value of feature $F$ present/absent in hidden state component $h$ was likely to have a strong effect, positive or negative, on model singular verb conjugation behavior.

\subsubsection{Filtering}

We considered a feature $F_c$ important if $\text{score}_s(F_c, B)>0.2$ or $\text{score}_p(F_c, B)>0.2$. We considered an edge $F_c \rightarrow F_e$ important if $\text{score}_s(F_c, F_e)>0.2$ or $\text{score}_p(F_c, F_e)>0.2$ \textit{and} both $F_c$ and $F_e$ were important. For each edge that crossed token boundaries we annotated the strongest detected value feature $F_v$ found to mediate between $F_c \rightarrow F_{v} \rightarrow F_e$ relationship, regardless of its importance.

If there proved to be between 3 or more features $F_c$, $F_e$, and $F_{i1}$, $F_{i2} ...$ both a direct $F_c \rightarrow F_e$ edge and a number of indirect sequences  $F_c \rightarrow F_{i1} (\rightarrow F_{i2}...) \rightarrow F_e$, we eliminated the $F_c \rightarrow F_e$ edge if, after setting all of the intermediate $F_i$ values to their normal unpatched values after AP'ing $F_c$, the relationship $F_c \rightarrow F_e$ weakened beneath the importance threshold.

We considered an edge $F_c \rightarrow W_U$ important if's strength was at least 20\% that of the strongest detected $F_c \rightarrow W_U$ edge over $S$.

We considered an edge $W_E \rightarrow F_e$ important if $F_e$ was in the first MLP layer and $F_e$ was itself important. 

We considered an edge $F_c \rightarrow W_U$ important based on manual inspection of the top tokens yielded by $F_c \cdot W_U$, seeing if they were predominantly singular or plural verb tokens, instead of irrelevant tokens.

We considered an edge to \emph{promote} if both $\text{sign}(|F_c(s_s)|) = \text{sign}(|F_e(s_{p\rightarrow s}|))$ and $\text{sign}(|F_c(s_p)|) = \text{sign}(|F_e(s_{s\rightarrow p}|))$. We considered an edge to \emph{inhibit} if both $\text{sign}(|F_c(s_s)|) != \text{sign}(|F_e(s_{p\rightarrow s}|))$ and $\text{sign}(|F_c(s_p)|) != \text{sign}(|F_e(s_{s\rightarrow p}|))$. Under unusual circumstances, this pair of equations would disagree, most such \emph{ambiguous} edges tended to not be very important to model behavior.

\section{Results and Discussion}

\subsection{SAE quality verification}

\begin{figure}[H]
  \centering
  \includegraphics[width=0.99\textwidth]{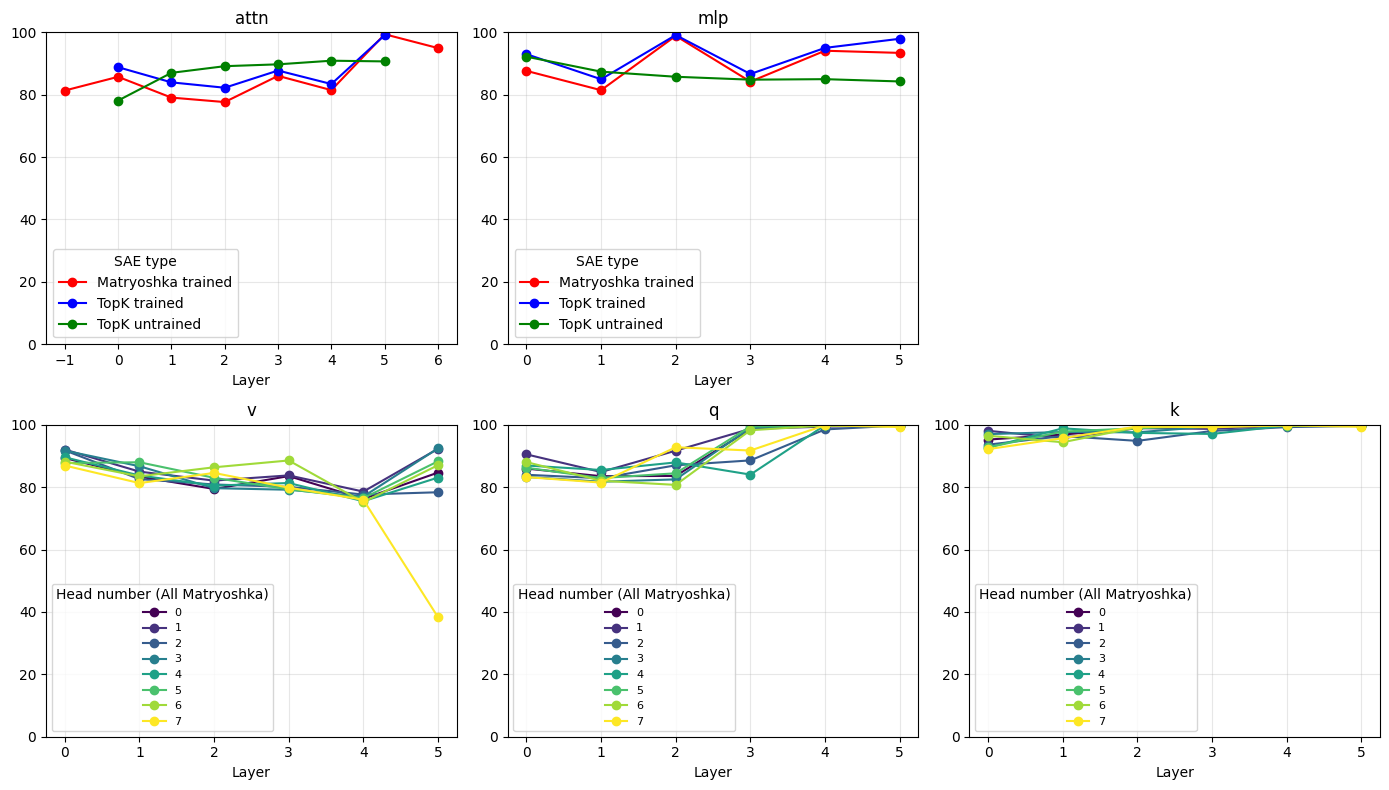}
  \caption{Variance explained by various SAEs on a calibration dataset of 1,000 hidden states. Our unique embedding and unembedding Matryoshka SAEs are for the sake of comparison denoted as ``Attention layer -1'' and ``Attention layer 6'' in the upper left subfigure. Our Gemma-2-2b and Llama-3.18b $W_E$ Matryoshka SAEs, not shown, had variance explained levels of 95.84\% and 94.31\%, respectively}
  \label{fig:SaeQuality}
\end{figure}

Using hidden states derived from the first 10,000 sentences in our SAE training dataset, we found that our SAEs were able to explain 80-90\% of the variance in the hidden states, depending on which model component in particular was being analyzed with only one exception - the final value component of the final attention head of the final layer (see Figure~\ref{fig:SaeQuality}. Matryoshka SAEs were similar but slightly inferior in performance on this metric, as compared to the TopK SAEs we trained on the attention and MLP components. This is line with the findings of Bussman et al. \cite{bussmann_learning_2025}, who introduced the Matryoshka SAE concept. Our embedding and unembedding matrix SAEs had similar variances explained as the comparable mlp and self-attention layer SAEs.

In addition, through analyzing the ability of the SAEs to interpret the untrained (specifically, checkpoint step 0) version of Pythia-70m, we found that statistical regularities in the distribution of tokens in the underlying corpus used to train an SAE are visible as regularities in the distributions of Pythia's corresponding hidden state vectors in both the trained and untrained models, and that training the Pythia model does not actually cause the hidden state vectors to become more predictable than they already were prior to training the model.

The vectors inside untrained Pythia's embeddings table are initialized using independent samples from the same Gaussian distribution. Thus, these vectors can be described in terms of the 512 uncorrelated Gaussian samples. Our TopK SAE, with $k=32$, creates a compressed description of the vector it attempts to reconstruct with a sequence of 32 $m, i$ pairs, with $m$ being feature magnitudes and $i$ being feature indices. If our TopK SAE had 512 features, we could create a maximum-efficiency reconstruction of an input vector $v$ by having the feature table consist entirely of 1-hot vectors and then encode the first 32 indices of the input vectors by setting the $m$ of each $m, i$ to the $i$th index of $v$. These representations would be able to represent 32/512 = 6.25\% of the variance of a large sample of vectors drawn from the same distribution as $v$. However, our SAEs instead have 8192 features each, 8 times the dimension of the vectors they are trying to reconstruct. Thus, an ideally constructed SAE would instead have a set of 8192 2-hot vectors covering 8/512 = 1.5625\% of all possible 2-hot vectors over a 512 dimensional basis. We could then sort all the samples $s_i \in v$ by size to generate pairs $s_a - s_b \approx 0$. We could subsequently capture two samples at once with a pair $m \approx s_a \approx s_b, i$ if there exists an $i$th two-hot feature vector for $a$ and $b$. The odds of this happening are 1.5625\%. If there is no suitable 2-hot vector, then we instead choose some feature $j$ with 2-hots at $a$ and $c$, set $m = s_a$, and accept a random reconstruction for sample $s_c$. Thus, each pair $m, i$ can explain about 1.015625 samples, placing an upper bound of 6.35\% variance explained by the SAE if the input vectors $v$ from our untrained Pythia are  distributed in a true  Gaussian fashion. The fact that our trained-Pythia-70m TopK SAEs and their untrained-Pythia-70m counterparts were able to explain about the same fraction of variance in the respective models' hidden state vectors suggests that the training process does not make the hidden state vectors more statistically regular but instead merely endows the hidden state vectors with a structure more amenable to language generation.

Finally, the SAEs trained on the Query and especially Key matrices were able to explain an exceptionally high proportion of the variance in the respective hidden state vector sets which they reconstructed. Despite this, this set of K and Q SAEs proved not useful for our purposes. For reasons which will be detailed below, we suspect this is due to these two model components having a much lower information density than the others, making SAEs, with their massive overcomplete feature spaces, an inappropriate tool to analyze them.

\subsection{Verb conjugation circuits identified in Pythia-70m}

\begin{figure}[!htbp]
  \centering
  \includegraphics[height=0.9\textheight]{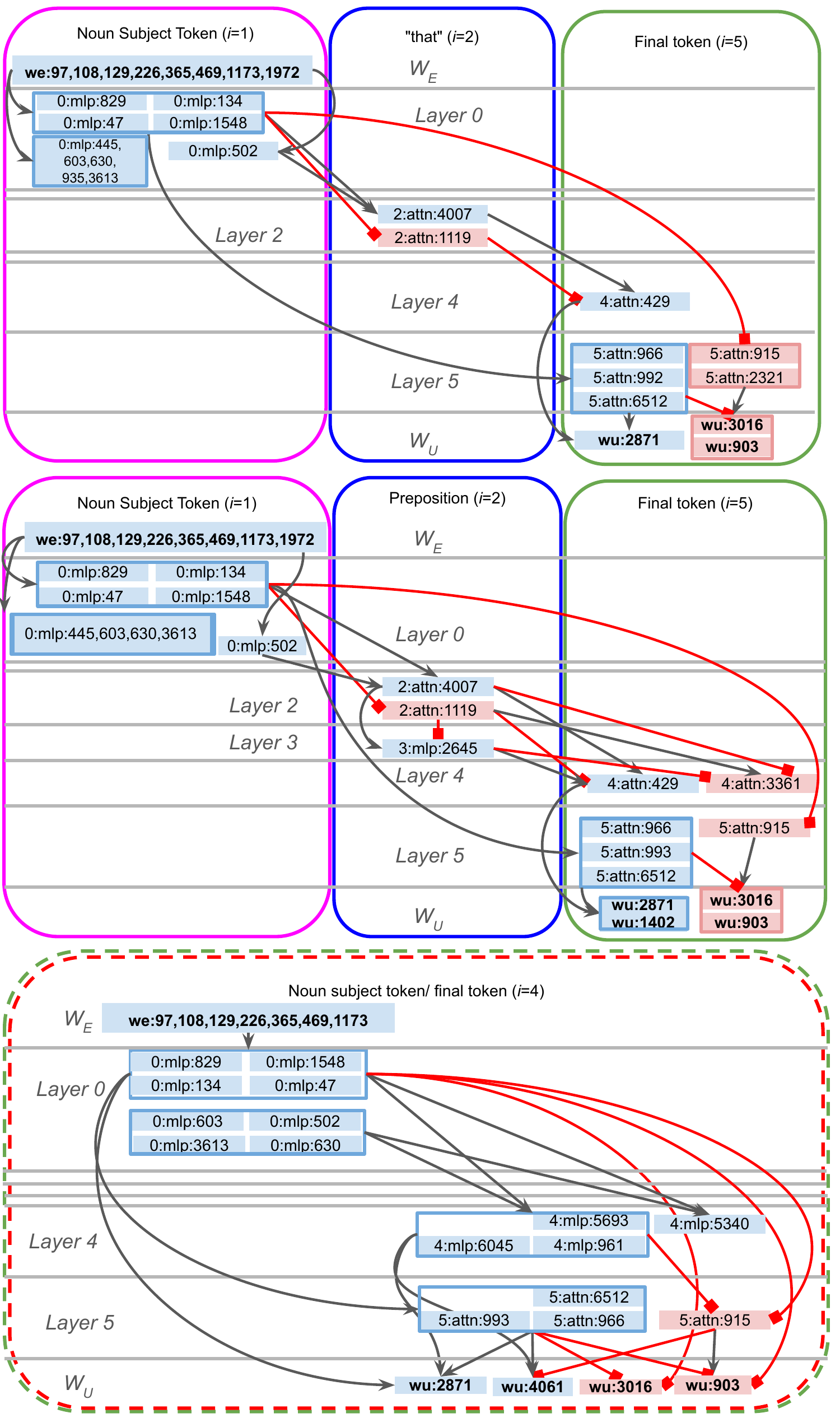}
  \caption{Verb conjugation circuits identified in \texttt{rc}, \texttt{nounpp}, and \texttt{within\_rc}. Sentences and verb continuations shown are merely examples, these circuits were derived by aggregating data from all sentence pairs and verb pairs in their respective datasets as described in the methods section. Single-high features are red, plural-high are blue. Black edges activate, red edges inhibit. The horizontal lines represent the layers of the LLM and the vertical blocks represent tokens in the prompt. Features belonging to the same sublayer and token which shared more than 60\% of ingoing and outgoing edges were grouped into boxes. }
  \label{fig:AllSAECircuits}
\end{figure}

Figure \ref{fig:AllSAECircuits} shows the verb conjugation circuits we identified in our  \texttt{rc}, \texttt{nounpp}, and \texttt{within\_rc} datasets. When we found large groups of features that exhibited almost identical edges in and out, we grouped them in red or blue boxes. The circuit we identified for \texttt{long\_rc} was too complex to show in a figure. 

With the exception of one of the datasets (\texttt{long\_rc}), we in fact did not find any hypothesized singular noun features of the kind we discussed above in the Methods section. We only found plural noun features, which we took to be an optimistic sign that our inductive methodology was not so overpowered as to be unfalsifiable. What we did find instead was that the majority of subject noun number features surfaced by our analyses were plural-high singular-low. Only circuit analysis of the \texttt{long\_rc} corpus identified any singular-high noun features. Because the SAE training procedure is designed to avoid learning features which would have negative values on any of the vectors of the training hidden states corpus (though in practice negative feature values occur with appreciable frequency, as is visible in Figure \ref{fig:context_free_number_features}), we interpret this as meaning that the grammatical singular may be the ``unmarked" noun number for Pythia-70m. This finding replicated in larger LLMs, discussed further in the next subsection.

Plural marking on nouns was unified in all scenarios. Though our different datasets surfaced various noun number features, there was heavy overlap between them, the feature we:1173 consistently had a strong effect size in all 4 datasets. We collected the tokens which most highly activated by this plural noun feature and some other prominent ones which interested us in Figure~\ref{fig:mlp-word-sets}.

The most important verb number choice selection mechanism in all datasets was plural noun number, through one circuit or another suppression of unembedding matrix feature wu:3016, which by default selects singular verbs. There happened to exist plural verb promotion features as well, but plural promotion was not as important as singular suppression.  Our analysis was not suited to identifying what it is that causes wu:3016 to be produced by default, as this would require us to compare verb-production-appropriate prompts to nonverb-production-appropriate prompts, giving us insight as to how Pythia-70m parses its prompts but massively increasing the scope of our analyses beyond what could fit in this paper.

The long distance agreement datasets surfaced 2 conjugation subcircuits, the "1-hop" circuit, which went directly from the subject token to the final token, which made use of the common verb production features, and the "2-hop" circuit, which made use of the contrasting +singular 2:attn:1119 and +plural 2:attn:4007 features, which competed to promote or inhibit the unique plural verb selection feature 4:attn:429. We found an interesting ``double negative'' sequence, present in each of the \texttt{rc}, \texttt{long\_rc}, and \texttt{nounpp} datasets, of the form plural noun ${\color{red}\rightarrow}$  2:attn:1119 ${\color{red}\rightarrow}$ 4:attn:429 $\rightarrow$ plural verb. This sort of ``double negative'' phenomenon is often found in biological interaction networks. This circuit appears to suppress plural verb production by default. 

The ``0-hop'' circuit used by \texttt{within\_rc}, which does not require the self-attention mechanism to function, activated a mixture of the common verb production features and some  verb production features in sublayer 4:mlp unique to this situation.

We found no 3-hop or greater circuits in any of the datasets, nor any 2-hop circuits that hopped onto any token other than the one immediately following the subject noun. Even though \texttt{long\_rc} analysis surfaced an enormous number of important features, all of them were found on the noun subject token, the token after it (`` that''), or on the final token.

\subsection{Privative Marked plural/Default singular noun number representation in Pythia-70m, Gemma-2-2b, and Llama-3.1-8b}

\begin{table}[ht]
    \centering
    \begin{tabular}{c|c|c|c|c}
        \hline
        $S$ LLM & Dataset & Feature & $\text{strength}_s(F_c, B)$& $\text{strength}_p(F_c, F_e)$ \\
        \hline
        Gemma-2-2b & \texttt{rc} & we:533 & 0.56 & 0.26 \\
        '' & '' & we:18 & 0.34 & 0.03 \\
        '' & '' & we:2645 & 0.22 & 0.01 \\
        '' & \texttt{nounpp} & we:533 & 0.47 & 0.25 \\
        '' & \texttt{within\_rc} & \textit{we:533} & 1.10 & 0.37 \\
        '' & '' & \textit{we:18} & 0.64 & -0.04 \\
        \hline
        
        Llama-3.1-8b & \texttt{rc} & we:440 & 0.46 & 0.69 \\
        '' & '' & \textit{we:936} & 0.21 & 0.15 \\
        '' & \texttt{nounpp} & we:440 & 0.33 & 0.67 \\
        '' & '' & \textit{we:936} & 0.21 & 0.18 \\
        '' & \texttt{within\_rc} & \textit{we:440} & 0.88 & 0.96 \\
        '' & '' & \textit{we:1121} & 0.33 & 0.32 \\
        '' & '' & \textit{\textcolor{red}{we:5548}} & 0.32 & 0.32 \\
        \hline

        Pythia-70m & \texttt{rc} & we:97 & 0.64 & 0.21 \\
        '' & '' & we:365 & 0.55 & 0.11 \\
        '' & '' & we:1173 & 0.53 & 0.13 \\
        '' & '' & we:108 & 0.53 & 0.11 \\
        '' & '' & we:469 & 0.45 & 0.06 \\
        '' & '' & we:129 & 0.40 & 0.06 \\
        '' & '' & we:226 & 0.30 & 0.03 \\
        '' & '' & we:1972 & 0.20 & 0.08 \\

        '' & \texttt{nounpp} & we:97 & 0.63 & 0.26 \\
        '' & '' & we:108 & 0.55 & 0.15 \\
        '' & '' & we:365 & 0.51 & 0.12 \\
        '' & '' & we:1173 & 0.50 & 0.12 \\
        '' & '' & we:469 & 0.44 & 0.09 \\
        '' & '' & we:129 & 0.37 & 0.08 \\
        '' & '' & we:226 & 0.33 & 0.05 \\
        '' & '' & we:1972 & 0.20 & 0.06 \\

        '' & \texttt{within\_rc} & we:97 & 0.74 & 0.19 \\
        '' & '' & we:1173 & 0.72 & 0.10 \\
        '' & '' & we:365 & 0.62 & 0.07 \\
        '' & '' & we:108 & 0.56 & 0.04 \\
        '' & '' & we:129 & 0.42 & 0.02 \\
        '' & '' & we:469 & 0.38 & 0.05 \\
        '' & '' & we:226 & 0.26 & 0.01 \\

        \hline
    \end{tabular}
    \caption{Encoding layer Matryoshka SAE features of all three LLMs. The feature we:5548 highlighted in red is the only singular positive feature we found.}
    \label{tab:strength-s}
\end{table}
After finding that our method yielded only plural-high singular-low SAE features in the embedding layer of Pythia-70m on all datasets except \texttt{long\_rc}, we extended our investigation to larger LLMs. We trained Matryoshka SAEs on the embedding matrices of these larger LLMs using identical methodology, training a 32768-feature SAE on Llama-3.1-8B's $W_E$ and a 16384-feature SAE on Gemma-2-2B's $W_E$. We gathered all features for which $\text{strength}_s(F_c, B) > 0.2$ or $\text{strength}_p(F_c, B)>0.2$ (it turned out that the latter condition never proved decisive, as we found that $\text{strength}_s(F_c, B) > \text{strength}_p(F_c, B)$ in most instances.) First, we discovered that our gradient based prefiltering methodology failed to identify important features in the embedding layer of the larger models, particularly in the case of \texttt{within\_rc}. So we instead brute-force calculated the values of $\text{strength}_s(F_c, B)$ and $\text{strength}_p(F_c, B)$ for all 16384 Gemma $W_E$ features and 32768 Llama $W_E$ features, which succeeded in identifying additional important number features for each, shown in italics in table \ref{tab:strength-s}. We also did these brute-force calculations for Pythia's $W_E$ SAE, but this found no features which we had not found earlier with our more computationally frugal methods. We found that all of these LLMs predominantly used privative plural/default singular marking, with only one singular positive feature found, we:5548 for the Llama x \texttt{within\_rc} experiment. In addition, we found that the larger models, despite being searched with larger SAEs, had \textit{fewer} verb-conjugation relevant number features, each model exhibiting a distinct dominant number feature (we:440 for Llama and we:533 for Gemma) across all 3 datasets with some auxiliary number features that only appeared in one or two of the datasets. This simplification-with-scale finding gave us cause for optimism that this method will scale well to even larger LLMs. In addition, we broadly found that singularizing a prompt by removing a plural marking feature had stronger effects than vice-versa for most Gemma and Pythia SAE features, though this was not true for Llama.

\subsection{Semantic fields in noun plural features}

We found that the different noun plural features identified in our searches had distinct semantics, as shown in figure \ref{fig:context_free_number_features}. we:108 is most prominent animate plural nouns, particularly those referring to people, like `` men'', `` women'', `` children'', `` people'', and `` babies'', in addition to many occupations. Many of we:1173's highly marked nouns were concrete inanimate objects.  we:365 has a rather peculiar ``religious/artistic'' plural noun semantic field that includes nouns related to art, learning, faith and communication. Feature wu:3016 appears to index plural nouns clustered around social and economic roles.

\begin{figure}[!htbp]
\centering
\setlength{\tabcolsep}{3pt}

\newtcolorbox{wordsetbox}[1]{
  enhanced,
  colback=gray!4,
  colframe=gray!55,
  colbacktitle=gray!15,
  coltitle=black,
  fonttitle=\bfseries\small,
  title={#1},
  boxrule=0.45pt,
  arc=1.5pt,
  left=4pt,
  right=4pt,
  top=3pt,
  bottom=3pt,
  before skip=0pt,
  after skip=0pt
}

\begin{minipage}[t]{\textwidth}
\begin{wordsetbox}{\texttt{we:1173}}
\RaggedRight\fontsize{10}{10}\selectfont
balloons  lamps  kernels  ghosts  clones -heads  trucks  masks  cameras  tables  sheets  fibers  blades  brushes  nodes  electrodes  gels  flares -boats -ICs  balls  folds  phones  beads  tags  sticks  monomers  wrists  buckets  arches  hooks -berries  poles  clusters  cards  qubits  locks  bubbles  triangles  flaps  pipes  bulbs  patches  bees  batteries  diodes  gates -istors  flies -Blocks  pits  lines  knots  capacitors  bearings  flows  grapes  screws  jets  fists  bundles  cores  roses  transistors  mortgages  trains  sails  chips  pods  shells  switches  cords  disks  cones  boxes  belts  pumps  detectors -balls  rods  roofs  tiles  mirrors  clocks  edges  floats  voltages  blocks  buses  loops  beams  grids  anchors  bottles  discs  LEDs  wires  pins -bows  pulses
\end{wordsetbox}
\end{minipage}
\hfill
\begin{minipage}[t]{\textwidth}
\begin{wordsetbox}{\texttt{we:365}}
\RaggedRight\fontsize{10}{10}\selectfont
Angels  tweets  Davies  youths  jokes  photographers  museums  GPs  students  emails -stars  commentators  textbooks  conversations  Music  texts  journals  feelings  Richards  ghosts  evenings  teachers  Blues  sessions  Friends  messages  lyrics  scholars  patrons  Schools  universities  listeners  thoughts  believers  concerts  Books  Africans  learners  historians  criticisms  Boys  painters  memories -friends  experiences  observers  dancers  prophets  perspectives -urches  songs  Ravens  Sessions  lovers  classics  visions  lectures  Eagles  magazines  Stories  speakers  manuscripts  workshops  notes  stories  celebrities  Christians  novels  musicians  intellectuals  Artists  singers  Songs  friendships  glasses  performers  composers  Catholics  classmates  Beatles  journalists  saints  sculptures  monks  poems  books  Fans  Girls  paintings  professors  followers  writers  Saints  blogs  writings  poets  friends  churches  fans  artists
\end{wordsetbox}
\end{minipage}

\vspace{5pt}

\begin{minipage}{\textwidth}
\begin{wordsetbox}{\texttt{we:108}}
\RaggedRight\fontsize{10}{10}\selectfont
makers  traders  hunters  manufacturers  criminals  attackers  writers  advocates  regulators  fishermen  performers  distributors  hospitals  buyers  suppliers  families  soldiers  crews  researchers  legislators  councils  athletes  fighters  corporations  departments  teenagers  teams  warriors  retailers -icians  engineers  designers  dealers  physicians  universities  startups  creators  girls  minorities  attendees  actors  districts  patients  organisations  riders  providers  volunteers  merchants  experts -people  members  cities  employers  counties  producers  women  villagers  teachers  patrons  participants  politicians  developers  practitioners  clients  dancers  organizers  miners  investors  artists  consumers  men  organizations  students  employees  customers  officials  inmates  agencies  entrepreneurs  players  firms  governments  residents  voters  workers  farmers  owners  businesses -ists  youths  managers  people  individuals  users  activists  executives  citizens  communities  leaders  companies
\end{wordsetbox}
\end{minipage}

\caption{The most strongly marked tokens for a number of prominent conjugation-involved plural noun features in the $W_E$ embedding matrix of Pythia-70m.}
\label{fig:mlp-word-sets}
\end{figure}

\subsection{The 2-hop circuit is responsible for a large portion of number distraction}

\begin{figure}[!htbp]
  \centering
  \includegraphics[width=0.7\textwidth]{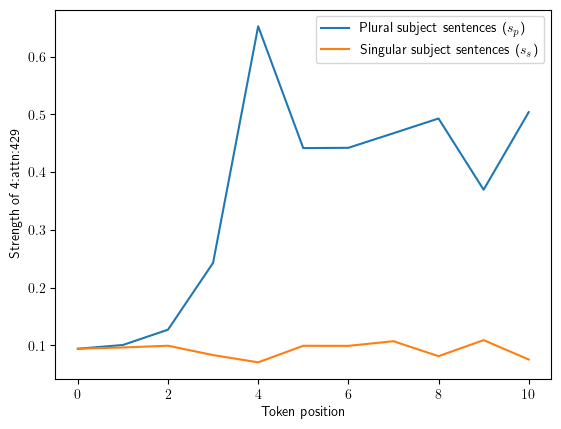}
  \caption{Average values of 4:attn:429 at different token positions in \texttt{long\_rc} The bird(s) that I saw eating the worms in the park this morning. blue - subject is plural; orange - subject is singular}
  \label{fig:Distraction4-attn-429}
\end{figure}

\begin{figure}[!htbp]
  \centering
  \includegraphics[width=0.7\textwidth]{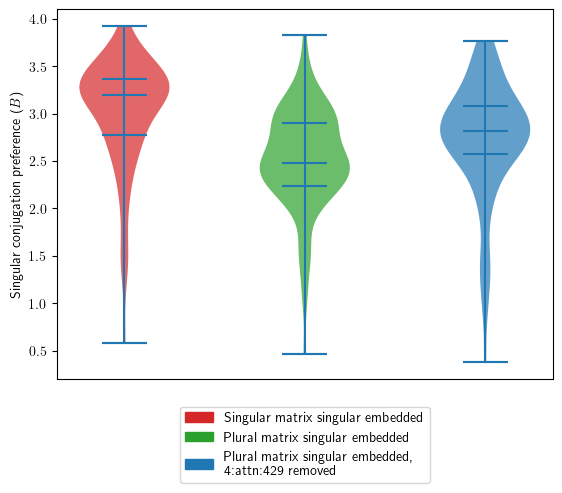}
  \caption{Removing 4:attn:429 from the final token of plural matrix singular embedded form sentences in \texttt{within\_rc} (i.e., ``The boys that the mother...'') increases singular verb conjugation preference}
  \label{fig:DistractionRemoval}
\end{figure}

We found that the 2-hop circuit, which predominantly promotes the plural verb production feature 4:attn:429 when the matrix subject is plural (symmetrical singular verb production features were visible in the analyses of \texttt{nounpp} and \texttt{long\_rc}, but their effects were weaker), appears to be the source of much of the number agreement distraction in \texttt{within\_rc}, as the two-hop circuit is mildly activated in it, clashing with the 0-hop circuit. The fact that the two-hop circuit is more effective at promoting plurality than vice-versa is likely the reason why number distraction is more severe in plural matrix singular embedded  sentences like ``The lawyers that the minister ...'', than vice versa, ``The lawyer that the ministers ...''. This finding is in agreement with observations related to neural correlates in the human brain of plural markedness in many languages \cite{Aleman2016, DUNAGAN2022105110}. The reason why the 0-hop and 2-hop circuits clash is because, as we show in data collected on \texttt{long\_rc} in \ref{fig:Distraction4-attn-429}, 4:attn:429 is copied onto all of the tokens of the embedded clause  when the matrix subject is plural (the embedded clause is every token from position 2 on, an example sentence in long\_rc is ``The birds| that he's out in the woods looking for''). Pythia-70m ``sprays and prays''  the plural number information, so to speak. This interferes with the model when it wants to produce a verb form which agrees with a singular noun inside the embedded clause, as happens in the \texttt{within\_rc} sentences. Removing it from plural matrix singular embedded sentences in \texttt{within\_rc}, as shown in figure \ref{fig:DistractionRemoval}, removes much of the distraction effect.

\subsection{Failure of SAEs to identify features in the Query and Key vectors}

We were unable to find features resembling our hypothesized \emph{Nsubj-$({\color{OliveGreen}W_Q^{(n,\ell)}}/{\color{blue}(W_K^{(n,\ell)}})$} feature. We picked out one of the most important attention head edges from our preceding analysis of the dataset \texttt{rc}, attention head 5 in layer 5 between tokens 1 and 5 (the subject and final tokens), which is responsible for propagating 56\% of the value of 5:attn:966 (the most prominent 'say a plural verb' feature) amd 0:mlp:134 (the most prominent 'plural noun' feature). We ablated one feature out of a time from ${\color{OliveGreen}q^{(5,5)}_5}$, the query vector responsible for this edge (the corresponding key vector is ${\color{blue}k^{(5,5)}_1}$, out of all 1024 features we had in our appropriate KQ Matryoshka SAE for Layer 5, head 5. We discovered that 11.04\% of all this SAE's features, when ablated, reduced the strength of the 0:mlp:134 $\rightarrow$ 5:attn:966 relationship by more than 50\%. We then preformed a control experiment using 1024 random feature ablation vectors instead (to simulate an untrained SAE). 11.62\% of all the control feature vectors also exhibited this effect. From this we immediately gathered that the contents of the KQ SAEs were not useful for examining attention edges, and abandoned further analysis of them.

\subsection{Verb form selection features are translinguistic}
The two most important verb production features we found across all datasets, resembling our hypothesized \textbf{Sg-verb-${\color{teal}W_O^{\ell}}$} and \textbf{Pl-verb-${\color{teal}W_O^{\ell}}$} features were wu:3016, which was singular-high plural-low, and  wu:2871, which was plural-high singular-low. We then assembled lists of the top 100 and top 1000 tokens whose unembedding matrix vectors had the highest dot product with these two feature vectors, and thus the two sets of tokens whose generation probabilities were most increased by their presence in $f$. The top-100 lists are shown in figure 
\ref{fig:say-verb-pair}. For the plural wu:2871, we found a number of foreign language tokens all meaning ``they are'' or ``they have'' in various romance languages, \texttt{sao} (Portuguese), \texttt{sont} (French), and \texttt{hanno} (Italian).

\begin{figure}[!htbp]
\centering
\newtcolorbox{attnwordsetbox}[1]{
  enhanced,
  colback=gray!4,
  colframe=gray!55,
  colbacktitle=gray!15,
  coltitle=black,
  fonttitle=\bfseries\small,
  title={#1},
  boxrule=0.45pt,
  arc=1.5pt,
  left=4pt,
  right=4pt,
  top=3pt,
  bottom=3pt,
  before skip=0pt,
  after skip=0pt
}

\begin{minipage}[t]{\textwidth}
\begin{attnwordsetbox}{\texttt{wu:2871}}
\RaggedRight\fontsize{10}{10}\selectfont
refer  generate  stimulate  communicate  specify  exceed  possess  represent  behave  evolve  disclose  dominate  describe  incorporate  contend  produce  give  conclude  consume  continue  find  achieve  adhere  characterize  see  define  reveal  think  operate  realize  don  coincide  propose  emphasize  hanno  believe  reside  develop  ARE  lose  undergo  exist  argue  feel  converge  suffer  emerge  utilize  determine  derive  haven  depict  accumulate  relate  originate  happen  resemble  contribute  know  make -uppose  deserve -regulate  HAVE  satisfy  constitute  correlate  contain  provide  illustrate  suggest 
-\textquotesingle re -\textquotesingle ve  comprise  arise -are  occur  demonstrate  include  owe  belong  imply  são  rely  indicate  remain  differ  require  vary -were  aren  involve  sont  appear  were -have  seem  do  are  have
\end{attnwordsetbox}
\end{minipage}
\hfill
\begin{minipage}[t]{\textwidth}
\begin{attnwordsetbox}{\texttt{wu:3016}}
\RaggedRight\fontsize{10}{10}\selectfont
delivers  arises  learns  pertains  mediates  realizes  aims  exceeds  calculates  understands  reveals  tells  evolves  extends  examines  illustrates  discloses  builds  isn  depends  improves  varies  gets  wants  relies  argues  celebrates  compares  represents  serves  helps  distinguishes  goes  maintains  manages -sembles  combines  brings  provides  spends  responds  demonstrates  adds  comes  regulates  recognizes  engages -ifies  explains  acknowledges  defines  seeks  interacts  enhances  introduces  encodes  differs  comprises  contributes  encourages  expands  reflects  utilizes  employs  predicts  contains  says  has  concludes  explores  Gets -includes  enjoys  speaks  grows  evaluates  emphasizes  participates  identifies  integrates  thinks  feels  believes  develops  describes  continues  considers  encompasses  relates  incorporates  agrees  promotes  involves  focuses  consists  proposes  discusses  teaches  includes  is
\end{attnwordsetbox}
\end{minipage}

\caption{Tokens selected by the most prominent Singular and plural verb production features in the $W_U$ unembedding matrix}
\label{fig:say-verb-pair}
\end{figure}

\subsection{Examining the ${\color{red}v}$ SAEs}
We were generally able to pinpoint large fractions of important $F_c \rightarrow F_e$ edges between attn and mlp features to a single  ${\color{red}v}$ feature inside of one attention heads (These would correspond to our hypothesized \textbf{Sg-verb-${\color{red}W_V^{(\ell,n)}}$} and \textbf{Pl-verb-${\color{red}W_V^{(\ell,n)}}$} features). In the case of the \texttt{rc} dataset, a single ${\color{red}v}$ feature was responsible for more than half of the AP effect strength in a given $F_c \rightarrow F_e$ for 53.7\% of the pluralization edges and 54.9\% of the singularization edges. However, the amount of information yielded from these searches was too overwhelming to be well-summarized in this article, and so future work to devise a strategy to tame the rat's nest of information our circuit-finding algorithm yields when allowed to inspect the individual attention heads would be valuable, particularly if we can also find an alternative means of decomposing the corresponding ${\color{OliveGreen}q}$ ${\color{blue}k}$ vectors into symbolic information.

\subsection{Alignment with observations in human language processing}

Our findings are consistent with the recently proposed collocation bootstrapping hypothesis in which statistical properties of the linguistic input to which children are exposed during language learning enable the acquisition of syntactic features including subject-verb agreement \cite{hobbs-mccoy-2026-collocational}. Similarly to the collocation bootstrapping whereby the child learns that what agrees with the verb is the noun in the subject position vs. the noun in the most recent position (e.g., the dogs in the park bark) by virtue of the words "dogs" and "bark" co-occurring more frequently than "park" and "barks" in the input, ANNs exposed to large volumes of presumed well-formed linguistic input would also encode these statistical properties, albeit in a very different manner than how a human brain encodes this information. The work showing that ANNs are capable of learning specifically subject-verb agreement information induced from training data dates back to well before modern LLMs \cite{elman1991}. The key to both human and machine learning processes appears to be not so much the learning mechanism but rather the distributional and frequency characteristics of the input from which they learn. 

The finding of the asymmetry in activation patching indicating that the grammatical singular may be the ``unmarked" noun number in all three LLMs used in the present work indicates that despite some differences in architecture, size, and training data, the models represent and propagate number agreement in a conceptually similar fashion. The notion of markedness of plural number in human languages dates back to Greenberg's universals, specifically Universal 33 "When number agreement between the noun and verb is suspended and the rule is based on order, the case is always one in which the verb precedes and the verb is in the singular" \cite{greenberg1963universals}. Multiple subsequent psycholinguistic \cite{BOCK1991,EBERHARD1997, Pearlmutter2000} and neurolinguistic \cite{Aleman2016, DUNAGAN2022105110} studies have confirmed this phenomenon. Our findings   are consistent with this prior work, particularly with human brain studies showing that plural nouns elicit greater cortical activity than singular nouns across multiple languages \cite{Aleman2016, DUNAGAN2022105110} and greater P600 event-related potential component in response to violation of plural vs. singluar subject-verb agreement \cite{Aleman2016, DUNAGAN2022105110}. We do not at all suggest that Transformer-based LLMs are a model for human brain (or any brain) beyond superficial similarities. The point of comparing our current work to prior work on number agreement processing in human brains is simply to highlight the fact that both ANNs and human brains learn this information from human language input and that the plural-singular asymmetry is likely to be the property of the input rather than an emergent property of the learning system.

\subsection{Limitations}

Because the SAE learns an overcomplete basis for the space of hidden state vectors it analyzes, many of these basis vectors are appreciably not-orthogonal with each other. Such features are mathematically confined to co-occurring (if the angle is acute) or not co-occurring (if the angle is obtuse). The nonlinear activation functions in the models' MLP layers are capable of allowing the model to distinguish two acute features, but we do not doubt that many of the features we found were redundant to some extent. To resolve this issue and trim down the number of SAE features under analysis would require devising a mathematical procedure for determining whether the weights in an MLP layer actually allow it to make a distinction between two acute-with-each-other feature vectors or not.

It is clear that certain features are shared between layers. In particular, all of the ``say-a-verb'' features we found, all features with strong direct connections to a $W_U$ feature $F_e$, with high $F_c \rightarrow F_e$ scores, were ``speaking the language'' of the unembedding matrix, and should have been identified as such, rather than identified as ``belonging to'' components in layer 4 or 5. Anthropic's recent interpretablility construct, the J-lens \cite{gurnee_verbalizable_2026}, helps determine which features rightfully belong to $W_U$, but it does not extend to $W_E$, nor to the even trickier middle-layer portions that use vector language wholly independent of the vocabulary.

In addition, we found "bundles" of noun number and verb number features that behaved almost identically. There were only insubstantial differences between the interactions that the plural noun features 0:mlp:134 and 0:mlp:829 had with any downstream features in any of the circuit diagrams, and similarly, 5:attn:966 and 5:attn:993, both being plural verb production features, were consistently coactivated in all contexts. We suspect that our SAEs spuriously split features that should have been reckoned as one, and thus greatly complicated our circuit diagrams. Further methodological developments in SAE training may solve this issue. One method we anticipate may help would be training the SAEs, not on the components of the residual streams themselves, but instead on the proposed updates which occur on the residual streams as the model trains. This may help isolate features on the residual streams which are actually significant to model behavior.

Our gradient based method for identifying number features worth further investigation failed on the larger models Llama-3.1-8B and Gemma-2-2B. The cause of this seemed to be gradient shrinking in the earlier layers. For our methods to scale to larger LLMs, it will be either necessary to massively increase compute (the feature searching process is very parallelizable) or devise an alternate means of cheaply identifying promising SAE features.

Though our methods could be extended with not too much difficulty to investigating changes in model behavior caused by multiple-token changes in the prompt (for example, by investigating sentence pairs like ``The dog in the park'' vs ``The two dogs in the park'', we currently do not have clear ideas about how our methodology could be used to investigate behaviors which require multiple tokens to manifest, even the ones as simple as how the model produces multi-token verbs (e.g., ``The judges uphold the law'' as opposed to ``The judges up|hold the law''. In our investigations we incidentally found that Pythia-70m appears to use the same features to produce the final token of multi-token singular verbs as it does those of monotokenal singular verbs, but we gleaned no insights as to what is going on when it produces the initial tokens of such verbs.

\clearpage

\bibliography{references}
\clearpage
\section{Acknowledgments}
The various versions of ChatGPT which were available to users of the \$20/month subscription tier 2024-2026 were used to assist in the authoring of code, primarily to assist in creating the figures. LLMs were \emph{not} used for generating prose.

The code we used for SAE training, analyses, and figure generation is available at \texttt{https://github.com/solin020/dissertation\_sae}.

\appendix

\section{Additional Figures}
\label{app:figures}
\begin{table}[H]
\centering
\caption{Weights}
\label{tab:weights}
\begin{tabular}{p{6cm}ll}
\toprule
Name & Variable & Dimension \\
\midrule
Token vocabulary & $V$ & $|V|$ \\
Embedding matrix & $W_E$ & $|V| \times d$ \\
Unembedding matrix   & $W_U$ &  $d \times |V|$\\
\midrule
Query matrix at attention sublayer $\ell$ in head $n$ & $W_Q^{(\ell,n)}$ & 
$d \times d_{h}$\\
Query bias at `` & $b_Q^{(\ell,n)}$ & 
$d_{h}$\\
Key matrix at `` & $W_K^{(\ell,n)}$ & 
$d \times d_{h}$\\
Key bias at `` & $b_K^{(\ell,n)}$ & 
$d_{h}$\\
Value matrix at `` & $W_V^{(\ell,n)}$ & 
$d \times d_{h}$\\
Value bias at `` & $b_V^{(\ell,n)}$ & 
$d_{h}$\\
LayerNorm scale for attention at `` & $\gamma_{KQV}^{(\ell,n)}$ & 
$d$\\
LayerNorm bias for attention at `` & $\beta_{KQV}^{(\ell,n)}$ & 
$d$\\
Attention output matrix at attention sublayer $\ell$ & $W_O^{\ell}$ & 
$d \times d$\\
Attention output bias at `` & $b_O^{\ell}$ & 
$d$\\
\midrule
MLP matrix 1 at sublayer $\ell$ & $W_\text{M1}^{\ell}$ & 
$d \times d_{i}$\\
MLP bias 1 at `` & $b_\text{M1}^{\ell}$ & 
$d_{i}$\\
MLP matrix 2 at ``& $W_\text{M2}^{\ell}$ & 
$d_{i} \times d$\\
MLP bias 2 at `` & $b_\text{M2}^{\ell}$ & 
$d$\\
LayerNorm scale for MLP at `` & $\gamma_{M}^{(\ell,n)}$ & 
$d$\\
LayerNorm bias for MLP at `` & $\beta_{M}^{(\ell,n)}$ & 
$d$\\
\midrule
\multicolumn{3}{c}{Corresponding SAE Weights} \\
\midrule
SAE Encoder matrix for attention sublayer $\ell$ output &
$\color{teal}{\text{SAE}(O^{\ell})_{W_e}}$ & $d \times d \cdot f$\\
SAE Encoder bias for `` &
$\color{teal}{\text{SAE}(O^{\ell})_{b_e}}$ & $d$\\
SAE Decoder matrix for `` &
$\color{teal}{\text{SAE}(O^{\ell})_{W_d}}$ & $d \cdot f \times d$\\
SAE Decoder bias for `` &
$\color{teal}{\text{SAE}(O^{\ell})_{b_d}}$ & $d$\\
SAE Encoder matrix for MLP sublayer $\ell$ output &
$\color{brown}{\text{SAE}(\text{M2}^{\ell})_{W_e}}$ & $d \times d \cdot f$\\
etc... & & \\
SAE Encoder matrix for query at sublayer $\ell$ head $n$ &
$\color{OliveGreen}{\text{SAE}(\text{Q}^{(\ell,n)})_{W_e}}$ & $d_h \times d_h \cdot f$ \\
etc... & & \\
SAE Encoder matrix for key  at ``&
$\color{blue}{\text{SAE}(\text{K}^{(\ell,n)})_{W_e}}$ & $d_h \times d_h \cdot f$ \\
etc... & & \\
SAE Encoder matrix for value at ``&
$\color{red}{\text{SAE}(\text{V}^{(\ell,n)})_{W_e}}$ & $d_h \times d_h \cdot f$ \\
etc... & & \\
xth feature vector for SAE for $\ell$th attention layer  & $\color{violet}{SAE(O^\ell)_x} =  \color{violet}{SAE(O^\ell)_{W_d}[x,:]}$& $d$ \\
etc... & & \\

\bottomrule
\end{tabular}
\end{table}

\begin{table}[htbp!]
\centering
\caption{States}
\label{tab:states}
\begin{tabular}{p{6cm}ll}
\toprule
Name & Variable & Dimension \\
\midrule
Prompt Token indices & $i$ or $j$ & $\text{prompt length}$ \\ 
Initial embeddings for token at index $j$ or for token `` man'' & $\color{magenta}{r_j^0}$ or $\color{magenta}{r_\text{`` man''}^0}$ & $d$ \\
Residual stream for token $j$/`` man'' after layer $\ell$  & $\color{magenta}{r_j^{\ell}}$ or $\color{magenta}{r_\text{`` man''}^{\ell}}$ & $d$ \\
Normed attention layer input for layer $\ell$ & ${\color{orange}r_{j}^{{\ell-1}:attnnorm}}$ or ${\color{orange}r_\text{`` man''}^{{\ell-1}:attnnorm}}$& $d$ \\
Key for '' at attention head $n$ of layer $\ell$ & $\color{blue}{k_j^{(n, \ell)}}$ or $\color{blue}{k_\text{`` man''}^{(n, \ell)}}$ & $d_h$ \\
Query for ''  & $\color{OliveGreen}{q_j^{(n, \ell)}}$ or $\color{OliveGreen}{q_\text{`` man''}^{(n, \ell)}}$ & $d_h$ \\
Value for ''  & $\color{red}{v_j^{(n, \ell)}}$ or $\color{red}{v_\text{`` man''}^{(n, \ell)}}$ & $d_h$ \\
Output for '' & $\color{teal}{o_j^{(n, \ell)}}$ or $\color{teal}{o_\text{`` man''}^{(n, \ell)}}$ & $d_h$ \\
Attention sublayer output for token $j$ or `` man'' for layer $\ell$ & $\color{violet}{O_j^{ \ell}}$ or $\color{violet}{O_\text{`` man''}^{\ell}}$ & $d$ \\
Attention paid by token $j$ to token $i$ at head $\ell$,$n$ & 
$a_{j \rightarrow i}^{(\ell, n)}$ or $a_{\text{`` yesterday''} \rightarrow \text{`` man''}}^{(\ell, n)}$& 1 \\
Normed mlp layer input for layer $\ell$ & ${\color{Peach}r_{j}^{{\ell-1}:mlpnorm}}$ or ${\color{Peach}r_\text{`` man''}^{{\ell-1}:mlpnorm}}$& $d$ \\
MLP sublayer $\ell$ output for token $j$ & $\color{brown}{m_j^{\ell}}$ or $\color{brown}{m_\text{`` man''}^{\ell}}$ & $d$ \\
The final residual stream vector of the last token & $\color{magenta}{f}$ or $\color{magenta}{r_{|s|}^{l}}$ & $d$\\

\bottomrule
\end{tabular}
\end{table}

\begin{table}[htbp!]
\centering
\caption{Dimensions}
\label{tab:dimensions}
\begin{tabular}{p{3cm}llp{3cm}}
\toprule
Name & Index Variable & Size variable & Size in Pythia-70m \\
\midrule
Vocab size & $t$ & $|V|$ & 50304 \\
Hidden size &  & $d$ & 512\\
Number of layers & $\ell$ & $l$ & 6 \\
Number of attention heads & $n$ & $n_h$ & 8 \\
Attention head size &  & $d_h = d / n_h$ & $64 = 512 / 8$ \\
MLP Intermediate size &  & $d_m$ & 2048 \\
Prompt length & $i$ or $j$ or `` token'' & $\text{prompt length}$ &  \\
\midrule
\multicolumn{4}{c}{Corresponding SAE dimensions} \\
\midrule
Feature / hidden size ratio & N/A & $f$ & 32 \\
Number of features & $x$ & $d \cdot f$ & 16384 (for sublayer outputs)\\
`` & $x$  & $d_h \cdot f$ & 2048 (for attention heads) \\
\bottomrule
\end{tabular}
\end{table}
\clearpage
\begin{figure}[htbp]
  \centering
  \includegraphics[width=\textwidth,height=\textheight,keepaspectratio]{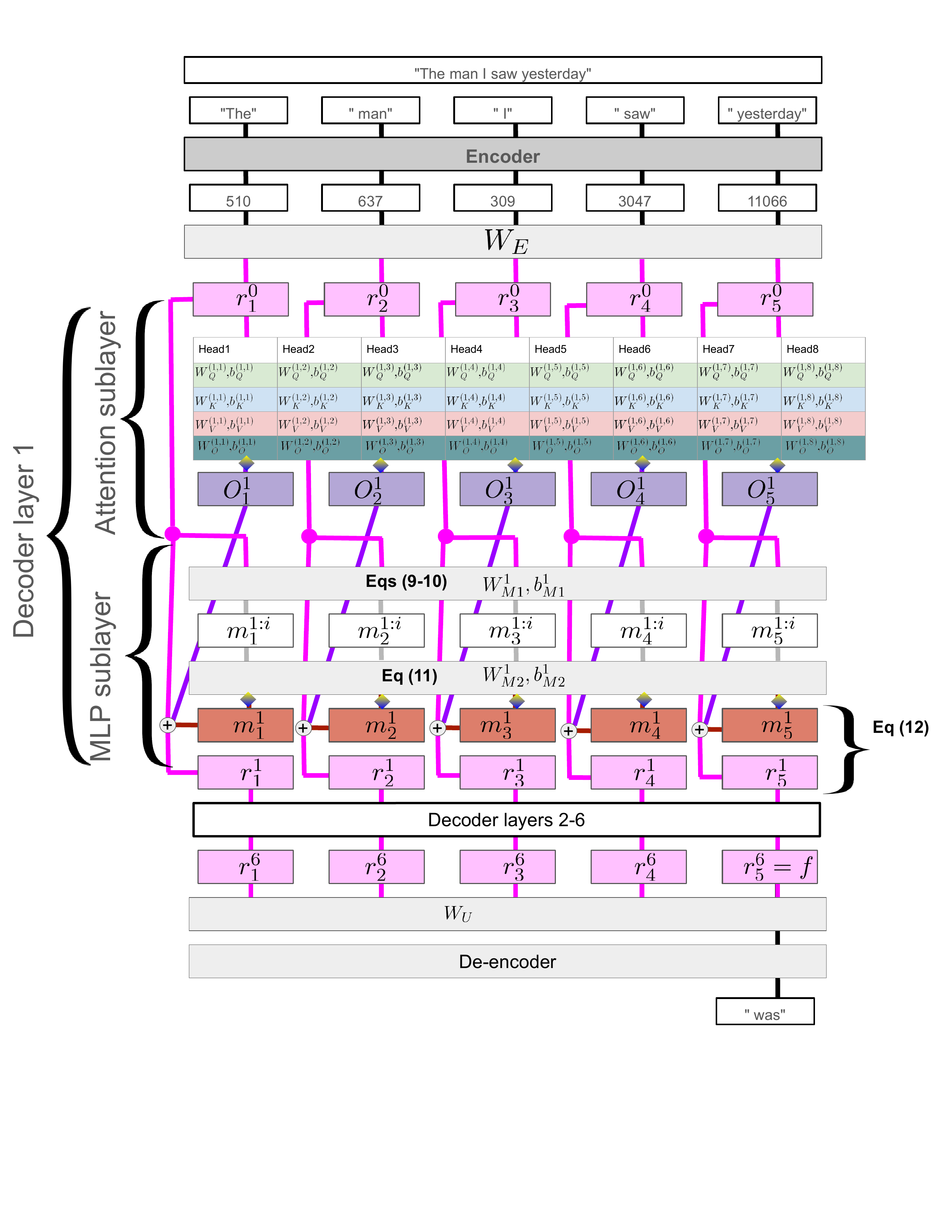}
  \caption{Schematic of Pythia-70m}
  \label{fig:LLMTopLevel}
\end{figure}
\clearpage
\begin{figure}[htbp]
  \centering
  \includegraphics[width=\textwidth,height=\textheight,keepaspectratio]{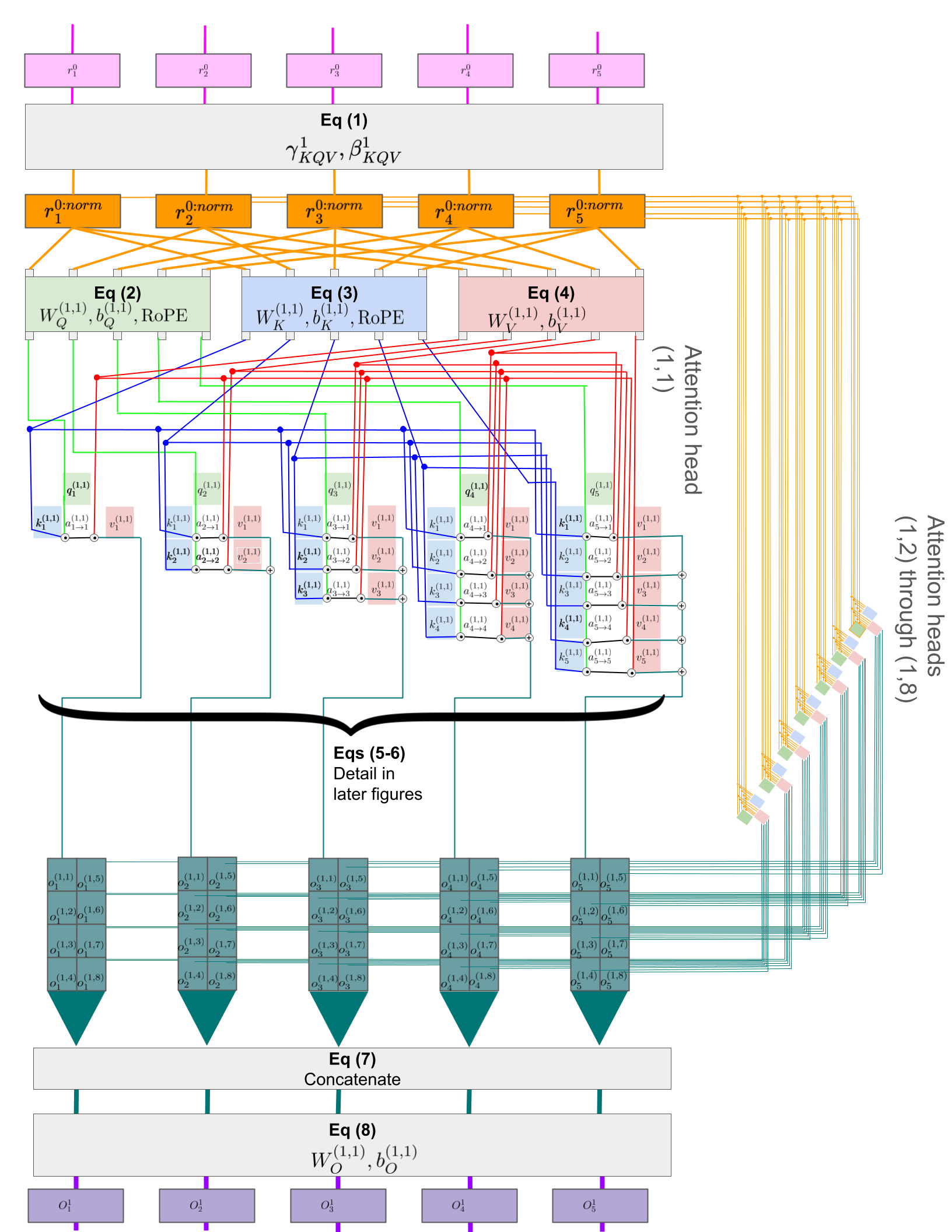}
  \caption{The 1st attention sublayer  of Pythia-70m}
  \label{fig:LLMAttentionFocus}
\end{figure}

\end{document}